\documentclass{article}

\usepackage{microtype}
\usepackage{graphicx}
\usepackage{booktabs}

\usepackage[accepted]{icml2024}

\usepackage{amsmath}
\usepackage{amssymb}
\usepackage{mathtools}
\usepackage{amsthm}

\usepackage[utf8]{inputenc}
\usepackage[T1]{fontenc}
\usepackage{url}
\usepackage[hidelinks]{hyperref}

\usepackage{booktabs} 
\usepackage{amsfonts} 
\usepackage{nicefrac} 
\usepackage{microtype}
\usepackage{lipsum}
\usepackage{graphicx}
\usepackage{caption} 
\usepackage[normalem]{ulem}
\usepackage{arydshln}
\usepackage{enumitem}
\usepackage{graphicx}
\usepackage{subcaption} 
\usepackage{varwidth}
\usepackage{changepage}
\useunder{\uline}{\ul}{}
\graphicspath{{media/}}

\newcommand{\methodname}{RoSA}

\usepackage{amsmath, amssymb}
\usepackage{algorithm}

\newcommand{\e}[1]{10^{#1}}
\newcommand{\bW}{\boldsymbol{W}}
\newcommand{\bX}{\boldsymbol{X}}
\newcommand{\bM}{\boldsymbol{M}}
\newcommand{\bO}{\boldsymbol{O}}
\newcommand{\bA}{\boldsymbol{A}}
\newcommand{\bB}{\boldsymbol{B}}

\newcommand{\bD}{\boldsymbol{\Delta}}
\newcommand{\wbar}{\bar{w}}
\newcommand{\dLd}[1]{\frac{\partial \f{L}}{\partial #1}}
\newcommand{\dbar}{\bar{\delta}}
\newcommand{\f}[1]{\mathcal{#1}}
\newcommand{\R}[2]{\mathbb{R}^{#1 \times #2}}

\DeclareMathOperator*{\argmin}{arg\,min}

\renewcommand{\paragraph}[1]{ \noindent \textbf{#1}}

\usepackage[capitalize,noabbrev]{cleveref}

\theoremstyle{plain}

\theoremstyle{definition}

\theoremstyle{remark}

\usepackage[textsize=tiny]{todonotes}

\icmltitlerunning{RoSA: Accurate Parameter-Efficient Fine-Tuning via Robust Adaptation}

\begin{document}

\twocolumn[
\icmltitle{RoSA: Accurate Parameter-Efficient Fine-Tuning via Robust Adaptation}

\icmlsetsymbol{equal}{*}

\begin{icmlauthorlist}
\icmlauthor{Mahdi Nikdan}{equal,ist}
\icmlauthor{Soroush Tabesh}{equal,ist}
\icmlauthor{Elvir Crnčević}{ist,tugraz}
\icmlauthor{Dan Alistarh}{ist,nm}
\end{icmlauthorlist}

\icmlaffiliation{ist}{ISTAustria}
\icmlaffiliation{nm}{Neural Magic}
\icmlaffiliation{tugraz}{Graz University of Technology}

\icmlcorrespondingauthor{Mahdi Nikdan}{mahdi.nikdan@ista.ac.at}
\icmlcorrespondingauthor{Soroush Tabesh}{soroush.tabesh@ista.ac.at}
\icmlcorrespondingauthor{Dan Alistarh}{dan.alistarh@ista.ac.at}
\icmlcorrespondingauthor{Elvir Crnčević}{elvir.crncevic@ista.ac.at}

\icmlkeywords{Machine Learning, ICML}

\vskip 0.3in
]
\printAffiliationsAndNotice{\icmlEqualContribution}

\begin{abstract}
    We investigate parameter-efficient fine-tuning (PEFT) methods that can provide good accuracy under limited computational and memory budgets in the context of large language models (LLMs). 
    We present a new PEFT method called Robust Adaptation (RoSA) inspired by robust principal component analysis   that jointly trains \emph{low-rank} and \emph{highly-sparse} components on top of a set of fixed pretrained weights to  efficiently approximate the performance of a full-fine-tuning (FFT) solution. 
    Across a series of challenging generative tasks such as grade-school math and SQL query generation, which require fine-tuning for good performance, we show that RoSA outperforms 
    LoRA,  pure sparse fine-tuning, and alternative hybrid methods at the same parameter budget, and can even recover the performance of FFT on some tasks. 
    We provide system support for RoSA to complement the training algorithm, specifically in the form of sparse GPU kernels which enable memory- and computationally-efficient training, and show that it is also compatible with low-precision base weights, resulting in the first joint representation combining quantization, low-rank and sparse approximations.
    {Our code is available at \url{https://github.com/IST-DASLab/RoSA}.}
\end{abstract}

\vspace{-1em}
\section{Introduction}
\setcounter{footnote}{1}
The advances brought about by large language models (LLMs) come with very large computational and memory costs, especially for training such models from scratch. 
In this context, fine-tuning LLMs using  limited data has become an effective and popular approach to improve performance on specific tasks, e.g.~\cite{wei2021finetuned, ouyang2022training, wang2022self, liu2022few}, or adapt LLMs to better fit expected  user behavior~\cite{askell2021general,bai2022training}. 
Yet, full fine-tuning of all LLM parameters (FFT), can be extremely expensive, especially in terms of memory cost, rendering this process prohibitive.  

Parameter-Efficient Fine-Tuning (PEFT) methods address this issue by allowing users to  optimize only over a restricted set of parameters, relative to the original model. On the one hand, this allows partial accuracy recovery relative to FFT, at a fraction of its computational and memory cost. 
An extremely popular recent instance of PEFT in the context of LLMs is given by the Low-Rank Adaptation (LoRA) family of methods~\cite{hu2021lora}, which train low-rank ``adapter'' layers for a selection of the model layers. LoRA methods are based on the intuition that the fine-tuning updates of pre-trained LLMs have low ``intrinsic rank'' during specialization to a sub-task, which allow these updates to be well-approximated by adapters. 
Besides memory and computational cost reductions, low-rank adaptation also has the advantage of implicit regularization, which can lead to more stable training, and simplify hyper-parameter search. 

\begin{figure}[t]
  \centering
  \includegraphics[width=.65\columnwidth]{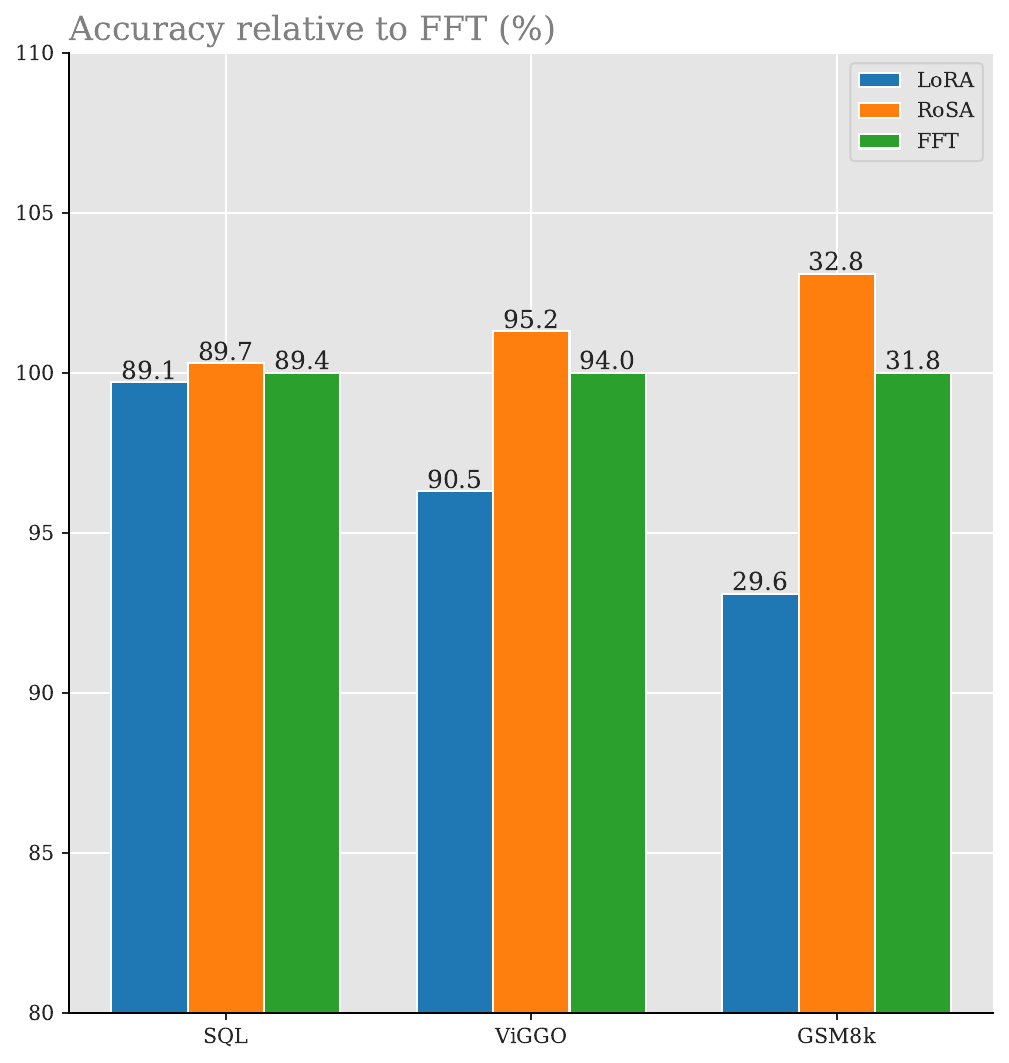}
  \caption{Comparison of the highest achieved accuracy by a single-epoch adaptation using various methods across three datasets on LLaMA2-7B, taken from our main experiments in Table \ref{table:main-results}. ({While LoRA and RoSA store parameters in \texttt{bfloat16} \cite{bfloat16} we use \texttt{float32} for FFT since they are more stable}). Each bar shows the percentage of accuracy relative to the accuracy achieved by FFT, and the numbers on top of the bars indicate the absolute accuracy.
  }
  \label{fig:barplot}
\end{figure}

One key weakness of LoRA-type methods is the fact that they can fail to recover accuracy for ``harder'' fine-tuning tasks, relative to FFT. This accuracy gap, illustrated in Figure~\ref{fig:barplot}, appears more likely to occur when the target tasks is more complex, such as the case for mathematical reasoning or coding tasks. 
It is therefore still an open question whether there exist PEFT methods which combine the good practical performance and ease-of-use of LoRA-type methods with the high accuracy of FFT. 

\paragraph{Contribution.} In this paper, we take a step towards addressing this question, by proposing a new PEFT method called \textbf{Ro}bu\textbf{S}t \textbf{A}daptation (\textbf{RoSA}). \methodname{} has similar computational and memory cost relative to LoRA-type methods, but is significantly more accurate at similar parameter and computational budgets, while being easy to use and tune. 
Specifically, in practical experiments \methodname{} essentially matches the accuracy of full fine-tuning, while offering stable convergence and relatively simple hyper-parameter tuning. We complement these algorithmic observations with a practical implementation, showing that \methodname{} preserves the memory advantage of LoRA-type methods. 

\begin{figure}[t]
  \centering
  \includegraphics[width=.9\columnwidth]{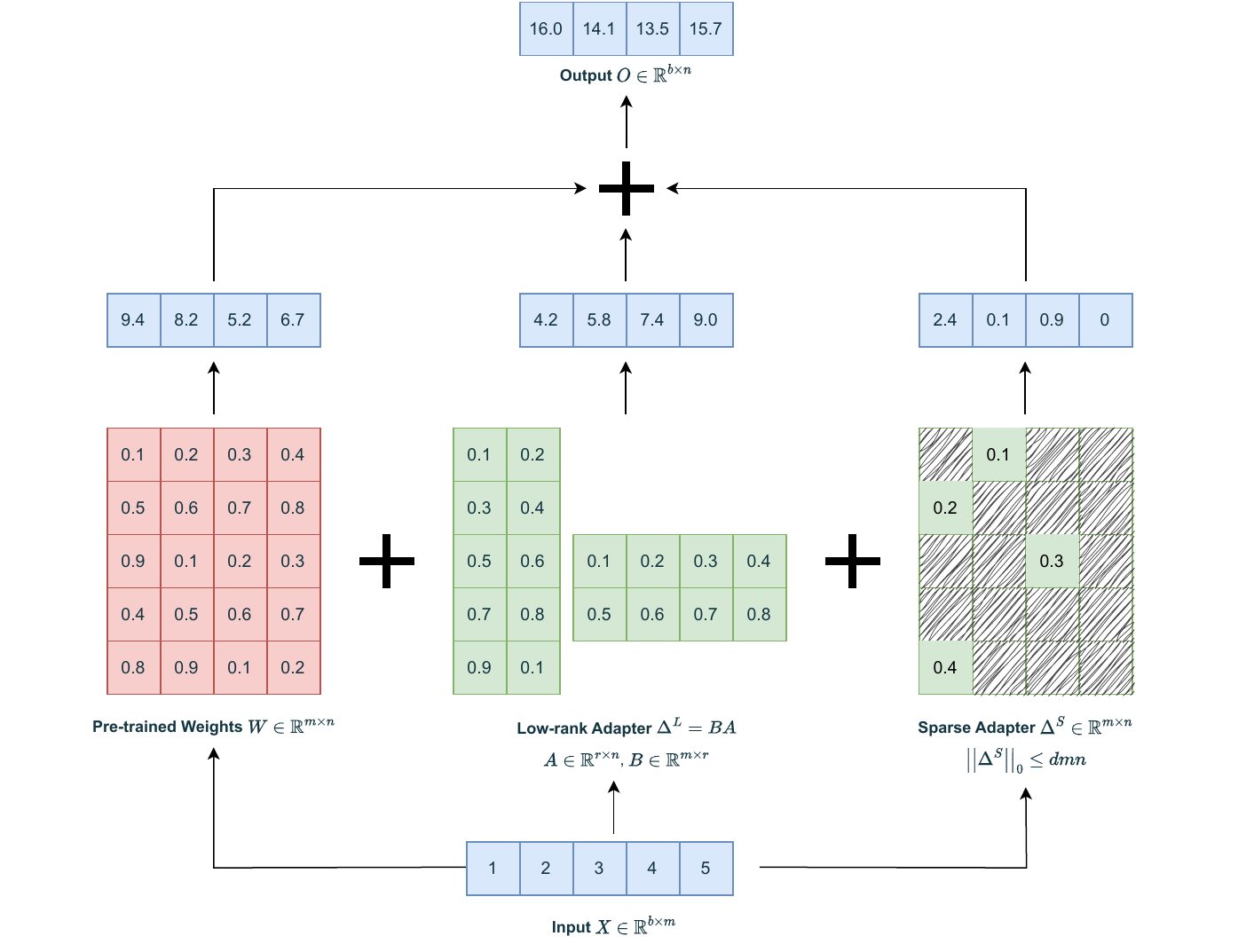}
  \caption{Illustration of Robust Adaptation (RoSA) applied to a single FC layer: In this instance, the weight matrix is of dimensions $5 \times 4$ and the batch size is $1$. The low-rank adapter has a rank of $2$, and the sparse adapter has a density of $20\%$. Trainable parameters are depicted in green, while red indicates parameters that remain frozen.}
  \label{fig:rosa}
\end{figure}

The motivation behind \methodname{} comes by revisiting the low ``intrinsic rank'' assumption that is the basis for the LoRA family of methods. Specifically, our investigation across several tasks shows that, while the FFT update can indeed be well approximated by a low-rank matrix, one can obtain a significantly better fit via a \emph{low-rank plus sparse matrix}, especially in the case of more complex tasks. 
Intuitively, the latter representation is better suited to matching outlier components which can cause a significant fraction of the compression error in the context of LLMs~\cite{dettmers2022llm, dettmers2023spqr}. 
This observation provides a connection to the area of robust principal component analysis (robust PCA)~\cite{candes2011robust}, which postulates that matrices arising from a noisy series of measurements can often be approximated as a sum between a low-rank component and a sparse one, and investigates algorithms for recovering such matrices. Starting from the hypothesis that the sum of gradient updates corresponding to FFT can be seen as an instance of robust PCA, we investigate methods for recovering such a sparse plus low-rank representation during training. 

Concretely, our proposed scheme trains \emph{two adapters}: a standard low-rank adapter, complemented by a sparse adapter, which are trained ``in parallel'' relative to the original pre-trained weights. The challenge is threefold, since we have to: 1) identify a highly-performant sparsity mask; 2) find a co-training mechanism which yields stable convergence; and, 3) provide  system support, specifically for an efficient sparse backward pass. 

Building on prior work in the area~\cite{sung2021training, chen2021dsee}, 
we resolve all three challenges and show that \methodname{} adapters can lead  to considerably higher accuracy of the resulting model, at a comparable parameter, memory, and computational budget relative to standard adapters that are either low-rank or sparse. 
We complement our algorithmic contribution with an efficient system implementation of  \methodname{} in Pytorch, that is fast on NVIDIA GPUs. 
Specifically, supporting sparse adapters with low  memory and computational overhead is non-trivial, as we must leverage sparse representations that are notoriously hard to support efficiently on GPUs~\cite{sputnik}. 

In addition, we extend our approach to support quantization of the base weights via QLoRA~\cite{dettmers2023qlora}, further improving efficiency at little or no accuracy cost. This results in a joint representation which recovers accuracy by combining all three common forms of compression: quantization, low-rank projections, and sparsity. 

In summary, we present promising evidence that the accuracy gap between adaptation methods and full fine-tuning of LLMs can be significantly reduced or even eliminated in some cases, without sacrificing practical accessibility. Therefore, \methodname{} can be an additional technique in the toolbox of machine learning practitioners working with LLMs in resource-constrained settings.

\vspace{-0.5em}
\section{Related Work}
\label{sec:related-work}

\paragraph{Parameter-Efficient Fine-Tuning.} Recent open LLMs~\cite{touvron2023llama, touvron2023llama2, opt, mpt} have demonstrated strong performance across various NLP tasks, but present challenges during training and inference due to high memory and computation cost. The common practice is to fine-tune these models on smaller downstream tasks rather than training from scratch \cite{min2021metaicl, wei2021finetuned, ouyang2022training, wang2022super, wang2022self, liu2022few}. While this approach partially addresses the computation demands, memory requirements are still a major concern. Parameter-Efficient Fine-Tuning (PEFT) methods have emerged as a solution  \cite{hu2021lora, adalora, li2021prefix, peft2, peft3, lester2021power, liu2022few, sanh2021multitask, peft7, krona, peft9, peft10, sung2021training}: Instead of fine-tuning all parameters, they selectively fine-tune smaller sets of parameters, potentially including a subset of the original ones. Notably, LoRA-type methods \cite{hu2021lora, adalora}, which train a low-rank perturbation to the original weights, have gained popularity for their efficiency and ease of use \cite{dettmers2023qlora}. However, it is known that they often fail to recover the accuracy of FFT \cite{krona, adalora}.

Earlier work focused on smaller-scale BERT-type models and sparse and/or low-rank updates. Specifically, FISH Mask \cite{sung2021training} updates only a sparse subset of weights in the BERT-base model \cite{devlin2018bert}. Its reliance on the Fisher Information Matrix (FIM) for generating sparsity masks renders it impractical for LLMs, unless heavy approximations are employed. FISH Mask uses the empirical diagonal estimation of the FIM. We examine its validity in Section \ref{sec:exp}, and find it to be less effective in the case of LLMs. 
{Relatedly, DSEE \cite{chen2021dsee}  trains a combination of low-rank and sparse adapters. However, despite promising results on BERT models, we find DSEE faces two main challenges in our setting. 
First, the DSEE sparsity masks perform a \emph{task-independent} decomposition of pre-trained weights. As we demonstrate in Section \ref{sec:exp}, this mask generation method does not effectively outperform random masks in the context of LLMs, and  significantly underperforms RoSA masks, even when applied to gradients instead of weights. Second, DSEE lacks system support for reducing costs by using a sparse adapter. In contrast,  RoSA comes with efficient GPU support, and is also compatible with weight quantization, as we show in QRoSA.}

\paragraph{Sparse Training / Fine-Tuning.}
Sparsity in language models has emerged as a popular strategy to address their significant computational and memory demands \cite{hoefler2021sparsity}, both for inference \cite{ gale2019state, singh2020woodfisher, sanh2020movement, frantar2022optimal} and training \cite{evci2020rigging, peste2021ac, hubara2021accelerated, jiang2022exposing, nikdan2023sparseprop}. A related research direction is sparse fine-tuning, where a network, pre-trained and sparsified on an upstream dataset, undergoes fine-tuning on a downstream task while keeping the sparsity mask fixed \cite{nikdan2023sparseprop, kurtic2022optimal, kurtic2023sparse}. Despite both sparse fine-tuning and sparse adaptation optimizing over a fixed subset of parameters, in sparse fine-tuning, the weights not involved are pruned (set to zero), whereas in sparse adaptation, they are merely frozen. This distinction allows us to achieve extremely high sparsity levels in sparse adaptation masks (over 99\%, see Section \ref{sec:exp}), whereas sparse training / fine-tuning typically struggles to 90-95\% without significant accuracy loss.

\paragraph{Robust Principal Component Analysis (RPCA).}
RPCA is a well-explored domain, focusing on techniques that can effectively handle data corrupted by outliers or gross errors. While classical Principal Component Analysis (PCA) assumes that the data is clean, RPCA methods extract robust principal components even in the presence of significant outliers \cite{rpcamultrim, rpcaransam,wright2009robust, candes2011robust, rpcainf1, rpcainf2, rpcaaltmin}. Specifically, given noisy measurements expressed as $A = L + S$, where $L$ is low-rank and $S$ is sparsely supported with elements of arbitrary large magnitude, the goal is to recover $L$ and $S$. While early approaches did not achieve this in polynomial time \cite{rpcainf1, rpcainf2, rpcaaltmin, rpcamultrim, rpcaransam}, recent papers show that it is possible to relax this by substituting the low-rank constraint on $L$ with a constraint on its nuclear norm~\cite{wright2009robust, candes2011robust}. 
By contrast, we perform Robust PCA-type optimization over a series of adapter matrices that are being learned jointly in an LLM. As such, existing theoretical mechanisms do not apply, although extending them would be an interesting question for future work. 

\paragraph{System Support for Sparsity.}
While PyTorch \cite{paszke2019pytorch} and STen~\cite{ivanov2022sten} have recently incorporated partial sparsity support for inference, obtaining benefits from unstructured sparse representations--as needed in our work--is notoriously challenging, especially on GPU hardware. 
So far, Sputnik~\cite{sputnik} is the only library to provide speedups in this context, although structured representations are known to be more amenable to speedups~\cite{openai, castro2023venom, magicube}. 
In this context, our kernels provide significant improvements upon Sputnik in the unstructured sparsity case by using a better indexing scheme and introducing a sparsity-adaptive SDDMM kernel for the backward pass.

\section{Adaptation of Large Language Models}

\subsection{Notation} 

Let $\f{N}$ represent a pre-trained Large Language Model (LLM), and let $\f{W} = \{\bW_1, \bW_2, ..., \bW_k\}$ denote a sequence of layers containing all fully connected weights of $\f{N}$, including sub-attention layers, with $\bW_i \in \R{m_i}{n_i}$ for all $1 \le i \le k$. Let the vector $\wbar \in \mathbb{R}^{\bar{d}}$ indicate the rest of $\f{N}$'s parameters (biases, normalization parameters, etc.) concatenated into a single vector. Given a dataset $\f{D}$ and a loss function $\f{L}(\f{D};\f{W}, \wbar)$, full fine-tuning (FFT) of $\f{N}$ on $\f{D}$ can be formulated as solving the  optimization problem:
\begin{equation}
\underset{\f{W}, \wbar}{\min} ~ \f{L}(\f{D}; \f{W}, \wbar)
\end{equation}
Given that LLMs typically contain billions of parameters, performing FFT can be slow and computationally expensive. This often renders it challenging or even impossible to execute on standard GPUs. A solution to this involves the application of adapters, which we will now formulate. Let $\Delta = \{\bD_1, \bD_2, ..., \bD_k\}$ include perturbations to the original fully connected weights, where $\bD_i \in \R{m_i}{n_i}$ for all $1 \le i \le k$. Define $\f{W}+{\Delta} = \{\bW_1+\bD_1, \bW_2+\bD_2, ..., \bW_k+\bD_k\}$. Additionally, let vector $\dbar \in \mathbb{R}^{\bar{d}}$ denote a perturbation to $\wbar$. The \textit{adapted} parameters are then found by solving the following optimization problem:
\begin{equation}
    \underset{\Delta, \dbar}{\min} ~ \f{L}(\f{D}; \f{W} + \Delta, \wbar + \dbar), ~~~~ \textit{s.t.} ~~~ \f{C}(\Delta, \dbar)
\end{equation}
where $\f{C}(\Delta, \dbar)$ is a set of constraints on the perturbations, such as low-rank or sparse, aiming to reduce the memory requirements or computational complexity of the optimization problem. Note that an adaptation with no constraints is equivalent to FFT.

In this context, our exclusive focus is on adaptations where $\dbar = \boldsymbol{0}$, as it aligns with standard practice. Nevertheless, given that $\wbar$ typically contains significantly fewer parameters than $\f{W}$, there is room for fine-tuning $\wbar$ as well. Also, we are specifically focusing on cases where all fully connected weights undergo adaptation, but our arguments extend trivially to the case where only a subset of these weights is being adapted. We now discuss a few special cases. 

\paragraph{LoRA: Low-Rank Adaptation.}
The well-known Low-Rank Adaptation (LoRA) \cite{hu2021lora} constrains the perturbations in $\Delta$ to exhibit a low rank, specifically the optimization objective will be the following:
\begin{equation}
    \begin{gathered}
        \underset{\Delta}{\min} ~ \f{L}(\f{D}; \f{W} + \Delta, \wbar), \\
        \textit{s.t.} ~~~ \forall ~ 1 \le i \le k:\textit{rank}(\bD_i) \le r
    \end{gathered}
\end{equation}
with $r$ being a fixed small number. This approach reduces the number of trainable weights for layer $i$ from $m_i n_i$ to $r (m_i + n_i)$, resulting in more memory-efficient fine-tuning. 

\paragraph{SpA: Sparse Adaptation.}
Sparse Adaptation (SpA), e.g. \cite{sung2021training}, imposes high sparsity constraints on perturbations, i.e., the optimization objective will be:
\begin{equation}
    \begin{gathered}
        \underset{\Delta}{\min} ~ \f{L}(\f{D}; \f{W} + \Delta, \wbar), \\
        \textit{s.t.} ~~~ \forall ~ 1 \le i \le k:||\bD_i||_0 \le d m_i n_i
    \end{gathered}
\end{equation}
where $d < 1$ represents the perturbation density and $||.||_0$ denotes the $\ell_0$ norm. It is common~(\citet{sung2021training, chen2021dsee}) to consider the case where each perturbation has a fixed support throughout training. This way, SpA reduces the number of trainable parameters by a factor  of $d$. 
At the same time, as discussed in Section~\ref{sec:related-work}, it encounters the primary challenges of  1) finding a good sparse support and 2) leveraging unstructured sparsity for speed and memory gains. 
Next, we discuss how our method approaches both challenges. 

\begin{algorithm}[tb]
 \footnotesize
   \caption{Mask Generation}
   \label{alg:mask}
\begin{algorithmic}
    \REQUIRE $\f{W}, \wbar \gets$ the fully connected weights and the rest of the LLM parameters, respectively
    \REQUIRE $\f{D_M} \gets$ the mask generation dataset, typically a small subset of the actual dataset
    \REQUIRE $\f{L}(.) \gets$ the loss function
    \REQUIRE $d \gets$ mask density
    \REQUIRE $\alpha \gets$ gradient accumulation exponent
    \STATE $\f{G} \gets \{\boldsymbol{0}, \boldsymbol{0}, ..., \boldsymbol{0}\}$
    \STATE $\texttt{[iterate through samples of $\f{D_M}$]}$
    \FOR {$s \in \f{D_M}$}
        \STATE $\texttt{[calculate the gradients for this sample]}$
        \STATE $\f{G}^s, \bar{g}^s \gets \nabla \f{L}(s; \f{W}, \wbar)$
        \STATE $\texttt{[accumulate the gradients]}$
        \STATE $\f{G} \gets \f{G} + (\f{G}^s)^{\alpha}$
    \ENDFOR
    \FOR {$G_i \in \f{G}$}
        \STATE $\texttt{[top-k elements of the accumulated grads]}$
        \STATE $\f{M}_i \gets \textit{TopK-Mask}(\f{G}_i,d \times \textit{numel}(\f{G}_i))$
    \ENDFOR
    \STATE \textbf{return} $\f{M}=\{\f{M}_1,\f{M}_2,...,\f{M}_k\}$
\end{algorithmic}
\end{algorithm}

\subsection{\methodname: Robust Adaptation}
We now describe our main adaptation method. 

\paragraph{\textbf{Motivation.}}
\label{sec:motiv}
One key drawback of existing LoRA-type methods is that, when faced with more complex downstream tasks, they often fail to match full fine-tuning accuracy (see Figure~\ref{fig:barplot}.) 
Intuitively, this occurs because the low-rank prior may not be able to capture the structure of more complex updates in this case, filtering important directions. 
 This filtering issue becomes particularly evident when conducting Singular Value Decomposition (SVD) on the FFT updates $\bD^*$ (defined as $\bD^* = \bW^{\text{FFT}} - \bW^{\text{BASE}}$) of LLM layers, as detailed in the Appendix \ref{apdx:fft_pca}. These analyses reveal that while $\bD^*$ is rank-deficient (see Figure \ref{fig:diff_svd}), it is not strictly low-rank. This distinction is characterized by the presence of a substantial fraction of singular values with relatively small, yet non-zero, magnitudes.

Robust Principal Component Analysis (RPCA) suggests an alternative in extracting \emph{robust} principal components via a low-rank matrix $L$ and a sparse matrix $S$. This decomposition offers a more nuanced approximation of the fine-tuning updates compared to solely low-rank methods.

\begin{figure}[ht]
\centering
\begin{subfigure}[b]{.49\columnwidth}
  \centering
  \includegraphics[width=\linewidth]{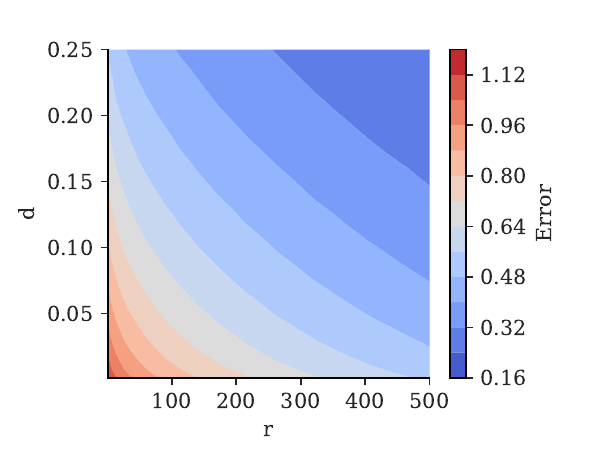}
  \caption{\label{fig:rpca_contour}}
\end{subfigure} %
\hspace{8pt}
\begin{subfigure}[b]{.45\columnwidth}
  \centering
  \includegraphics[width=\linewidth]{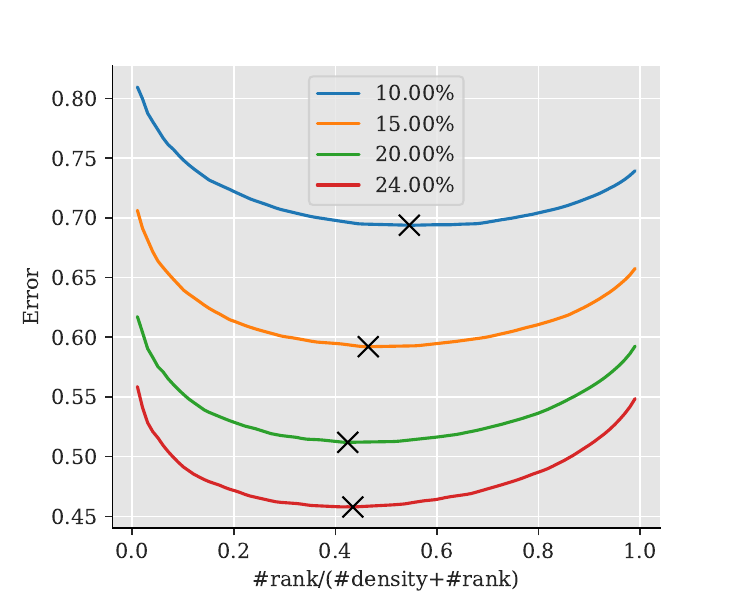}
  \caption{\label{fig:rpca_slice}}
\end{subfigure}
\caption{\label{fig:rpca} 
Illustration of the Frobenius norm error (Figure \ref{fig:rpca_contour}) of a Robust PCA approximation to the full-fine-tuning update, for an arbitrary layer (\texttt{l:20, v\_proj} of LLaMA2-7B, while varying rank and sparsity independently.  Figure \ref{fig:rpca_slice} depicts slices of Figure \ref{fig:rpca_contour} with similar parameter counts, showcasing the trade-off between sparsity and low-rank under different parameter budgets.
}
\end{figure}

To demonstrate the potential of using a combination of sparse and low-rank matrices to approximate a $\bD^*$ fine-tuning perturbation in the context of LLMs, we apply an RPCA solver to extract robust principal components $\tilde\bD^* = \tilde\bD^S + \tilde\bD^L$ of a randomly selected layer of LLaMA2-7B for a given sparsity and rank. In Figure \ref{fig:rpca_contour}, we have analyzed a randomly selected module from LLaMA2-7B, computed its $\bD^*$ when fine-tuned on the GSM8k dataset, and then applied GreBsmo RPCA solver \cite{grebsmo-zhou13b}, with varying ranks and densities for the low-rank and sparse components. The results in Figure \ref{fig:rpca_slice} clearly demonstrate that, given a parameter budget to approximate $\bD^*$, \emph{employing a combination of low-rank and sparse approximations yields a more accurate representation than using either approach in isolation}.

This analysis motivates our joint use of low-rank and sparse fine-tuning. The link between RPCA and RoSA lies in the former’s introduction of the low-rank and sparse decomposition, a concept we leverage in RoSA to enhance the efficiency and accuracy of fine-tuning LLMs. 
In practice, our approach will do this in a task-adaptive fashion by ``warming up'' a LoRA instance for a short training interval and then identifying the largest sparse directions for improvement. 

\paragraph{\textbf{Formulation.}} We formulate the optimization objective of Robust Adaptation (\methodname) as follows:
\begin{equation}
\begin{gathered}
    \underset{\Delta^L, \Delta^S}{\min} ~ \f{L}(\f{D}; \f{W} + \Delta^L + \Delta^S, \wbar), \\
    \textit{s.t.} ~~~ \forall ~ 1 \le i \le k: \left\{\begin{matrix}
    \vspace*{4pt}
    \textit{rank}(\bD_i^{L}) \le r \\ \vspace*{4pt}
    ||\bD_i^{S}||_0 \le d m_i n_i
    \end{matrix}\right.
\end{gathered}
\end{equation}
where $\Delta^{L}$ and $\Delta^{S}$ represent the low-rank and sparse adapters, respectively. In practice, we generate the sparsity masks using Algorithm \ref{alg:mask}, and then optimize the low-rank and sparse adapters jointly. Refer to Figure \ref{fig:rosa} and Appendix Algorithm \ref{alg:rosa}  for a detailed description of \methodname. 

\section{System Implementation}
\label{sec:sys}
In this section, we briefly describe our efficient implementation of \methodname, detailed in full in Appendix \ref{apdx:system}.

\paragraph{\textbf{Low-Rank Format.}} Similar to \citet{hu2021lora}, we store an $m \times n$ low-rank adapter with rank $r$ as the multiplication of two matrices $\bB \bA$, where $\bB$ and $\bA$ are $m \times r$ and $r \times n$, respectively.

\paragraph{\textbf{Sparse Format.}} Sparse adapters are stored in Compressed Sparse Row (CSR) format, which utilizes three lists to represent an $m \times n$ sparse matrix with $nnz$ non-zero values: a \texttt{values} list with size $nnz$, storing the non-zero values; a \texttt{row-offsets} list with size $m + 1$, indicating the position of the first non-zero element in each row within the \texttt{values} list; and a \texttt{column-indices} list with size $nnz$, containing the column index of each corresponding element in the \texttt{values} list. Additionally, in line with Sputnik~\cite{sputnik}, an extra \texttt{row-indices} list with size $m$ is included, sorting rows based on their non-zero element count. In our case, this \texttt{row-indices} list is employed for load-balancing and kernel launch configuration purposes.  

{\paragraph{\textbf{Forward Pass.}}
Consider a single fully connected layer with an adapted weight matrix $\bW + \bD^L + \bD^S$ of size $m \times n$. For simplicity, assume there is no bias vector. Given a batch of inputs $\bX$ of size $b \times m$, the layer output is expressed as:
\begin{equation}
    \label{eq:O}
    \begin{aligned}
        \bO &= \bX (\bW + \bD^L + \bD^S) \\
        &= \bX (\bW + \bD^S) + (\bX \bB^L)\bA^L
    \end{aligned}
\end{equation}
Calculating the term $\bW + \bD^S$ requires the addition of sparse and dense matrices, for which we provide an efficient kernel detailed in Appendix \ref{apdx:system}. It is worth noting that the multiplication in the second term is decomposed into two multiplications with low-rank, making it extremely fast.}

{\paragraph{\textbf{Backward Pass.}}
Given the gradients of the output $\dLd{\bO}$, the backward pass through a layer involves calculating the gradients of the parameters and inputs, as follows:  
\begin{equation}
    \label{eq:dLdX}
    \begin{aligned}
        \dLd{\bX} &= \dLd{\bO} (\bW + \bD^L + \bD^S)^T \\
        &= \dLd{\bO} (\bW + \bD^S)^T + \Big( \dLd{\bO} (\bA^L)^T \Big) (\bB^L)^T
    \end{aligned}
\end{equation}
\begin{equation}
    \label{eq:dLdBL}
    \begin{aligned}
        \dLd{\bB^L} = \dLd{(\bB^L\bA^L)} (A^L)^T = X^T \Big( \dLd{\bO} (A^L)^T \Big)
    \end{aligned}
\end{equation}
\begin{equation}
    \label{eq:dLdAL}
    \begin{aligned}
        \dLd{\bA^L} = (B^L)^T \dLd{(\bB^L\bA^L)} = \Big( (B^L)^T X^T \Big) \dLd{\bO}
    \end{aligned}
\end{equation}
\begin{equation}
    \label{eq:dLdDS}
    \dLd{\bD^S} = X^T \dLd{\bO}
\end{equation}
Similarly to formula \ref{eq:O}, Equations \ref{eq:dLdX}, \ref{eq:dLdBL}, and \ref{eq:dLdAL} can also be computed efficiently. However, the implementation of equation \ref{eq:dLdDS} has a specific structure called a Sampled Dense-Dense Matrix Multiplication (\texttt{SDDMM}) \cite{nikdan2023sparseprop}, i.e. multiplying two dense matrices where only specific elements of the output are needed.

\paragraph{Leveraging Mask Structure.} While general \texttt{SDDMM} is efficiently supported in e.g., \texttt{sputnik}, one special feature of our setting is that non-zero values in RoSA masks tend to cluster in a small subset of rows/columns, as illustrated in Appendix~\ref{apdx:system}. We suspect that this is correlated to the low-rank structure of the complementary adapter. To exploit this, we provide a new specialized \texttt{SDDMM} implementation which leverages this observation to maximize efficiency, specifically by dynamically skipping fully-zero rows and columns when present, depending on the specific sub-matrix structure. 
Compared to the SOTA  \texttt{sputnik} kernels, our RoSA kernel achieves a geometric mean speedup of 1.36x and a peak speedup of 3x on LLM matrices. We provide a full discussion of matrix structure, kernel descriptions and layer-wise speedups in Appendix \ref{apdx:system}.}

{
\paragraph{Gradient Accumulation.} As explained in Algorithm \ref{alg:mask}, creating the masks involves accumulating full gradients, which can be challenging in terms of memory. To address this, we adopt a simple solution by transferring the gradients of each weight matrix to CPU as soon as it is computed. This ensures that, at most, one weight matrix's gradient is stored on GPU at any given time. We note that this approach does not affect the runtime significantly, as the mask generation dataset is typically very small ($32$ samples in our experiments).
}

\begin{figure}[t]
  \centering
  \includegraphics[width=1\columnwidth]{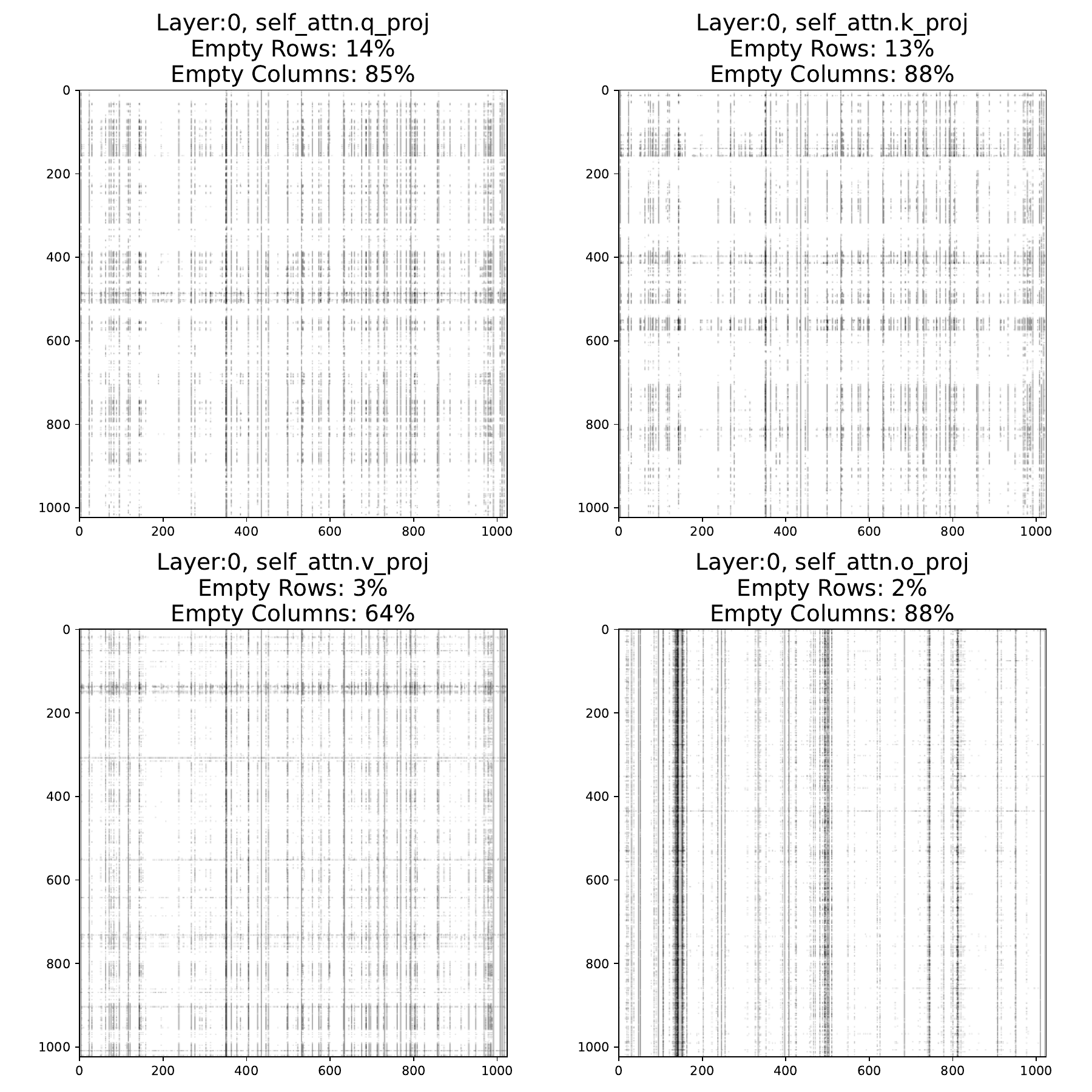}
  \caption{Illustration of row and column sparsity structure for the RoSA masks. Specifically, a subset of masks in the LLaMA2-7B model is visualized with a max-pool kernel of size 4 and stride 4, showing that a  fraction of around 50\% of the parameter rows and columns are completely zero. 
  }
  \label{fig:masks_subset}
\end{figure}

\section{Experiments}

\begin{table*}[t]
\centering
\caption{{Comparison of fine-tuning LLaMA2-7B using FFT, LoRA, SpA, and RoSA in terms of memory usage and accuracy on three datasets. For RoSA, we examine different splits of the parameter budget into low-rank and sparse adapters. ($\dagger$) Our experiments show that the single-epoch FFT results on ViGGO are suboptimal when the parameters are stored in \texttt{bfloat16}. Single-epoch \texttt{float32} FFT results on GSM8k, ViGGO, and SQL are $31.8$, $94.0$, and $89.4$, respectively.}}
\label{table:main-results}
\scalebox{0.9}{
\begin{tabular}{ccccccccc}
\toprule
                  &       &       &         \multicolumn{2}{c}{\texttt{GSM8k}}          & \multicolumn{2}{c}{\texttt{ViGGO}}          & \multicolumn{1}{c}{\texttt{SQL}}           \\ \cline{4-8}
                  & \texttt{\#Params} & \texttt{Memory} & \texttt{1 Epoch} & \texttt{Extended} & \texttt{1 Epoch} & \texttt{Extended} & \texttt{1 Epoch} \\ \toprule 
\texttt{FFT} & $6.7$ \texttt{B} & $>60$ \texttt{GB} & $\boldsymbol{32.3}$ & $\boldsymbol{38.8}$ & $\boldsymbol{82.1}$ & $\boldsymbol{95.0}$ & $\boldsymbol{89.0}$ \\
\toprule  
\texttt{LoRA} $r=16$ & $41.1$ \texttt{M} & $20.6$ \texttt{GB} & $28.4$ & $\boldsymbol{37.8}$ & $90.5$ & $95.8$ & $88.7$ \\
\texttt{RoSA} $r=12,d=0.15\%$ & $41.0$ \texttt{M} & $20.3$ \texttt{GB} & $\boldsymbol{31.2}$ & $36.0$ & $\boldsymbol{95.0}$ & $96.5$ & $88.3$ \\
\texttt{RoSA} $r=8,d=0.3\%$ & $40.8$ \texttt{M} & $20.3$ \texttt{GB} & $29.2$ & $37.5$ & $94.5$ & $\boldsymbol{97.1}$ & $77.6$ \\
\texttt{RoSA} $r=4,d=0.45\%$ & $40.6$ \texttt{M} & $20.3$ \texttt{GB} & $30.6$ & $35.5$ & $93.4$ & $96.6$ & $\boldsymbol{89.7}$ \\
\texttt{SpA} $d=0.6\%$ & $40.4$ \texttt{M} & $20.3$ \texttt{GB} & $26.2$ & $29.5$ & $72.6$ & $89.8$ & $83.2$ \\
\toprule 
\texttt{LoRA} $r=32$ & $82.3$ \texttt{M} & $20.9$ \texttt{GB} & $29.6$ & $36.2$ & $87.0$ & $96.8$ & $\boldsymbol{89.1}$ \\ 
\texttt{RoSA} $r=24,d=0.3\%$ & $81.9$ \texttt{M} & $20.6$ \texttt{GB} & $30.5$ & $37.8$ & $94.4$ & $95.8$ & $88.9$ \\
\texttt{RoSA} $r=16,d=0.6\%$ & $81.6$ \texttt{M} & $20.7$ \texttt{GB} & $\boldsymbol{32.2}$ & $\boldsymbol{38.6}$ & $\boldsymbol{95.2}$ & $\boldsymbol{97.1}$ & $88.3$ \\
\texttt{RoSA} $r=8,d=0.9\%$ & $81.2$ \texttt{M} & $20.7$ \texttt{GB} & $30.3$ & $37.2$ & $94.5$ & $96.9$ & $88.9$ \\
\texttt{SpA} $d=1.2\%$ & $80.9$ \texttt{M} & $20.7$ \texttt{GB} & $21.9$ & $29.9$ & $45.8$ & $95.7$ & $74.2$ \\
\toprule 
\texttt{LoRA} $r=64$ & $164.5$ \texttt{M} & $21.7$ \texttt{GB} & $27.4$ & $35.5$ & $76.9$ & $95.0$ & $88.7$ \\
\texttt{RoSA} $r=48,d=0.6\%$ & $163.8$ \texttt{M} & $21.3$ \texttt{GB} & $30.5$ & $38.2$ & $93.0$ & $96.6$ & $88.1$ \\
\texttt{RoSA} $r=32,d=1.2\%$ & $163.1$ \texttt{M} & $21.4$ \texttt{GB} & $32.2$ & $36.2$ & $93.4$ & $\boldsymbol{97.3}$ & $\boldsymbol{89.2}$ \\
\texttt{RoSA} $r=16,d=1.8\%$ & $162.4$ \texttt{M} & $21.5$ \texttt{GB} & $\boldsymbol{32.8}$ & $\boldsymbol{38.4}$ & $\boldsymbol{95.1}$ & $96.5$ & $84.6$ \\ 
\texttt{SpA} $d=2.4\%$ & $161.7$ \texttt{M} & $21.8$ \texttt{GB} & $29.6$ & $37.2$ & $92.3$ & $95.7$ & $87.8$ \\
\toprule
\end{tabular}
}
\end{table*}

We now provide experimental support for the effectiveness of \methodname, and of QRoSA, its variant with quantized base weights. The following subsection outlines the experiment settings, including details on the network and datasets. To ensure a fair comparison, we conducted thorough and careful tuning for each adaptation method, details of which are described next. We then present the results, along with ablation studies, showcasing the improvements achieved by \methodname. Finally, we also assess \methodname's memory utilization, highlighting that it requires the same resources as LoRA and SpA in a fixed parameter budget while offering significantly improved accuracy.

\label{sec:exp}
\subsection{Settings}
\paragraph{\textbf{Setup, Model and Datasets.}} 
We integrated RoSA into a fork of the standard \texttt{PEFT} library \cite{peftlibrary} and performed all the experiments using the MosaicML \texttt{llm-foundry} codebase \cite{llm-foundry}.
We perform fine-tuning of the LLaMA2-7B model \cite{touvron2023llama2} on three standard datasets: ViGGO \cite{viggo}, GSM8k \cite{gsm8k}, and SQL generation~\cite{Seq2SQL, yu2018spider}, containing $5.1k$, $7.47k$, and $30k$ training samples and $1.08k$, $1.32k$, and $1k$ test samples, respectively. Refer to Appendix \ref{sec:qual} for examples of the GSM8k dataset. In the case of SQL, we follow the dataset formation strategy described in~\cite{Anyscale}. On GSM8k, we only consider the accuracy of the final answer. Notably, these datasets are chosen such that they are highly specialized and, therefore, require fine-tuning for good performance: for example, on GSM8k, the pre-trained LLaMA-2 model has 0\% one-shot accuracy, and the multi-shot accuracy is also very poor (around 6\%). 

\paragraph{\textbf{Hyperparameters.}} In all experiments, we use a standard batch size of $32$ (micro-batch size $1$ + gradient accumulation) and a maximum context length of 512, which matches the dataset sample structure. We employ the AdamW optimizer \cite{loshchilov2017decoupled} with parameters $\beta_1=0.9$, $\beta_2=0.999$, $\epsilon=\e{-8}$, and a linear learning rate scheduler with 20 batches warmup. Notably, all floating-point values are stored in  \texttt{bfloat16} \cite{bfloat16}, popular due to low memory usage and good accuracy. Our main experiments run for a single epoch, but we demonstrate in ablation studies that extended training can further improve adaptation results. Following \cite{hu2021lora}, we use $\alpha=16$ and a dropout of $0.05$ for the low-rank adapter, while experimenting with various $r$ values ranging from $4$ to $64$. 
Additionally, we set the size of the mask generation dataset to 32 samples in all experiments while tuning the gradient accumulation exponent ($\alpha$ in Algorithm \ref{alg:mask}) as a binary hyperparameter ($1$ for averaging gradients and $2$ for diagonal Fisher).

The sparse adapter's density ranges from $0.15\%$ to $2.4\%$. While it is possible to adapt only a subset of the linear layers in the model, we specifically consider the case where every fully connected layer undergoes adaptation. This choice is motivated by the significantly lower memory usage of adaptation parameters compared to storing the original parameters (see Tables \ref{table:main-results} and \ref{tab:qrosa}). The best learning rates for single-epoch FFT are $4 \times \e{-5}$, $2\times \e{-5}$, and $1\times \e{-4}$ on SQL, ViGGO, and GSM8k, respectively, while for extended FFT it is $4 \times \e{-5}$ on ViGGO and $5 \times \e{-5}$ on GSM8k. For LoRA and SpA parameters, the best-performing learning rates are selected in the range $[\e{-4}, \e{-3}]$ and $[\e{-4}, 8 \times \e{-4}]$, respectively.
In RoSA experiments, we find it beneficial to initially fine-tune solely with LoRA for 64 batches, generate and fix the sparse masks, and restart training with both LoRA and sparse adaptation (SpA) activated. {All experiments, except for FFT, comfortably run on a single NVIDIA GeForce RTX 3090 GPU  $24.3$ GB memory (see Table \ref{table:main-results}).

\subsection{Results}
\paragraph{\textbf{Main Experiment.}} In Table \ref{table:main-results}, we summarize our main experiments, which examine the accuracy of various fine-tuning approaches at various budgets across all the tasks considered. We consider three parameter budgets: $40$ million, $80$ million, and $160$ million. For each budget, we explore five different ways of distributing parameters between LoRA and SpA, ranging from pure LoRA/SpA to intermediate sparse + low-rank budgets.  The main experiments are conducted for a standard single pass over the dataset (epoch). However, for the smaller ViGGO and GSM8k datasets, we observe that extended training improves adaptation results. Hence, we also present the best results for each method from 2 and 3 epochs on these two datasets under the `Extended` label. (We did not run extended training on SQL due to its much larger size.) {Additionally, for QRoSA, we follow~\citet{dettmers2023qlora} and report the accuracy of the single-epoch adaptations when the pre-trained weights are 4-bit double-quantized.}

\paragraph{Single-Pass Runs.} The results in Table  \ref{table:main-results} show that, across all tasks and budgets, RoSA outperforms both LoRA and SpA. The only exception is the $80M$ budget trained on SQL, where LoRA marginally outperforms RoSA ($89.1$ vs $88.9$). However, on the same task, RoSA $40M$ achieves a remarkable $89.7$ accuracy.
Surprisingly, in the single-epoch regime, RoSA even surpasses FFT significantly on all three datasets, highlighting the fast convergence of the hybrid adapter approach.
This shows that this approach can be particularly effective in the context of short, single-pass training, across tasks and parameter budgets. 

\paragraph{Extended Training Experiments.} 
The above conclusion still holds in extended experiments, where we find that RoSA can, in fact, match or even outperform FFT on both GSM8k ($38.6\%$ vs $38.8\%$) and ViGGO ($97.3\%$ vs $95.0\%$). Additionally, except for the $40M$ GSM8k, RoSA outperforms both LoRA and SpA. These results complement our single-pass experiments, indicating the superiority of RoSA in longer multiple-pass regimes. 
The fact that some of the best results for extended training are obtained on the medium-sized parameter budget suggests that the computational budget should be balanced against the active parameters for the run: the largest budget tends to yield the highest performance 
 on the larger SQL dataset.

Overall, these results clearly highlight the effectiveness of RoSA; specifically, we find it remarkable that we are able to fully recover FFT accuracy while using parameter budgets that are 40-100x smaller. 
Finally, the memory overheads of maintaining sparse and low-rank components are indeed low: all our experiments fit inside a single 24GB GPU.

{\paragraph{QRoSA: Quantizing Pre-trained Weights.}
Following QLoRA \cite{dettmers2023qlora}, we repeat the single-pass experiments while double-quantizing the pre-trained weights to total memory. We observe that QRoSA slightly lags behind QLoRA in the larger budgets on the SQL dataset. However, it outperforms every other method (including FFT) on GSM8k by achieving $33.1$ accuracy. Remarkably, in this setting, we need less than $12$ \texttt{GB} of memory to match or exceed the accuracy of FFT on LLaMa2-7B!}


\begin{table}[h]
\caption{{Comparison of fine-tuning LLaMA2-7B using different adaptation methods in terms of memory usage and accuracy on three datasets, while the pre-trained weights are 4-bit double-quantized following \citet{dettmers2023qlora}.}}
\label{tab:qrosa}
\scalebox{0.8}{
\begin{tabular}{ccccc}
\toprule
                  & \texttt{Memory} & \texttt{GSM8k} & \texttt{ViGGO} & \texttt{SQL} \\ \toprule 
\texttt{FFT} & $>60$ \texttt{GB} & $\boldsymbol{32.3}$ & $\boldsymbol{82.1}$ & $\boldsymbol{89.0}$ \\
\toprule  
\texttt{QLoRA} $r=16$ & $12.6$ \texttt{GB} & $29.8$ & $88.0$ & $88.2$ \\
\texttt{QRoSA} $r=12,d=0.15\%$ & $10.7$ \texttt{GB} & $\boldsymbol{31.8}$ & $93.8$ & $88.5$ \\
\texttt{QRoSA} $r=8,d=0.3\%$ & $10.7$ \texttt{GB} & $30.9$ & $\boldsymbol{95.0}$ & $\boldsymbol{88.6}$ \\
\texttt{QRoSA} $r=4,d=0.45\%$ & $10.7$ \texttt{GB} & $30.3$ & $92.4$ & $86.7$ \\
\texttt{QSpA} $d=0.6\%$ & $10.8$ \texttt{GB} & $22.8$ & $89.5$ & $79.2$ \\
\toprule
\texttt{QLoRA} $r=32$ & $13.0$ \texttt{GB} & $25.6$ & $74.7$ & $\boldsymbol{89.0}$ \\ 
\texttt{QRoSA} $r=24,d=0.3\%$ & $11.0$ \texttt{GB} & $30.4$ & $93.3$ & $88.3$ \\
\texttt{QRoSA} $r=16,d=0.6\%$ & $11.1$ \texttt{GB} & $\boldsymbol{33.1}$ & $93.8$ & $86.6$ \\
\texttt{QRoSA} $r=8,d=0.9\%$ & $11.1$ \texttt{GB} & $32.8$ & $\boldsymbol{95.4}$ & $83.7$ \\
\texttt{QSpA} $d=1.2\%$ & $11.3$ \texttt{GB} & $28.0$ & $93.0$ & $85.0$ \\
\toprule
\texttt{QLoRA} $r=64$ & $13.8$ \texttt{GB} & $30.6$ & $88.1$ & $\boldsymbol{89.4}$ \\
\texttt{QRoSA} $r=48,d=0.6\%$ & $11.9$ \texttt{GB} & $30.5$ & $93.6$ & $81.6$ \\
\texttt{QRoSA} $r=32,d=1.2\%$ & $11.9$ \texttt{GB} & $\boldsymbol{32.3}$ & $94.3$ & $88.2$ \\
\texttt{QRoSA} $r=16,d=1.8\%$ & $12.0$ \texttt{GB} & $30.8$ & $\boldsymbol{95.0}$ & $88.5$ \\ 
\texttt{QSpA} $d=2.4\%$ & $12.2$ \texttt{GB} & $28.9$ & $90.8$ & $42.9$ \\
\toprule
\end{tabular}
}
\end{table}

{
\paragraph{Hyper-parameter Selection.} Given a parameter budget, RoSA introduces a new hyper-parameter: the ratio by which we distribute the budget between the sparse and low-rank components. Our results in Table \ref{table:main-results} show that in many cases there is a threshold for the LoRA rank above which the results do not improve further. The existence of this rank threshold was already known before, e.g., Section 7.2 in the original LoRA paper \cite{hu2021lora}. In our experiments, this is more nuanced on the GSM8k and ViGGO datasets, where the optimal rank across different budgets is around 12-16, and the rest of the budget should be assigned to the sparse component to achieve the best results. This is justified by the fact that the difference between FFT and pre-trained weights has only a few large singular values (Figure \ref{fig:diff_svd}). On the other hand, while hyper-parameter tuning is required to achieve the best results, we found that in almost all cases simply distributing the budget equally between the low-rank and sparse adapters is enough to outperform other adaptation methods. Hence distributing the budget half-half can serve as a solid default choice.
}

 \paragraph{\textbf{Mask Choice Ablation.}} We investigate the impact of different mask generation methods of RoSA for the GSM8k dataset in Table \ref{tab:masks}. Let $\tau_{d}(\cdot)$ be the TopK magnitude mask with density $d$. 
 Then the methods we consider are:
 
 \vspace{-0.3cm}
 \begin{enumerate}
 \setlength\itemsep{0em}
     \item \textit{GradMag-LW (ours)}: 
     $\f{M}=\tau_{d}(\nabla \bW^{+\tilde L})$\\
     A TopK magnitude mask on the accumulated square of gradients as described in Algorithm~\ref{alg:mask} following warm-up of the low-rank instance, where $\bW^{+\tilde L} := \bW+\tilde\bD_L$ and $\tilde\bD_L$ is the partially-trained low-rank instance.
     \item \textit{GradMag/GradFish}: $\f{M}=\tau_{d}(\nabla \bW)$\\
     A TopK magnitude mask on gradients accumulated at initialization (in $\ell_1$ or $\ell_2$ norm squared), following FISH Mask \cite{sung2021training}.
     \item \textit{WeightRPCA}: $\f{M}=\tau_{d}(\bW_S)$\\
     The sparse component resulting from RPCA on the weights ${\bW}$, $\bW_S$, with a target density of $d$, following DSEE \cite{chen2021dsee}.
    \item \textit{GradRPCA}: $\f{M}=\tau_{d}(\nabla \bW_S)$\\
     The sparse component resulting from RPCA on the weight gradient $\nabla{\bW}$, $\nabla \bW_S$, with a target density of $d$, which we see as a natural combination of FISH Mask and DSEE. 
     \item \textit{Lottery Ticket Update Masking (LTM)}: $\f{M}=\tau_{d}(\bD^{*}_{S})$\\
     For this, we try to identify a good set of coordinates to optimize over ``in hindsight'', by computing the sparse component of RPCA over the FFT update $\bD^*$, denoted by $\bD^{*}_{S}$, with a target density of $d$.
     \item \textit{RND($d$)}: A random mask with density $d$.
 \end{enumerate}

\begin{table}[h]
\centering

\caption{\label{tab:masks} Comparison of various masking methods: Training of LLaMa2-7B Model on GSM8k for 1 epoch using 80M trainable parameters.}
\begin{tabular}{lc}
\toprule
\textbf{}          & \textbf{GSM8k}    \\ \cline{2-2}
\textbf{Method}    & \textbf{Accuracy} \\ \hline
LTM                & \textbf{33.66}             \\ \hline
GradMag-LW \small{(ours)}         & \textbf{32.16}             \\
GradMag (FISH Mask)           & 30.10             \\
GradRPCA           & 29.87             \\
WeightRPCA (DSEE)         & 30.71             \\
RND                & 30.25             \\ \toprule
\end{tabular}
\end{table}

First, we observe that the ``Lottery Ticket'' Mask (LTM), which has hindsight knowledge of the best optimization directions from the perspective of the FFT update, predictably performs very well, being in fact competitive with FFT accuracy on GSM8k. The second best-performing method, by a significant margin, is given by the RoSA masks, coming within $\sim$ 1\%  of the ideal mask. The remaining methods essentially perform within the variance of choosing random initial masks. The fact that gradient RPCA at initialization significantly under-performs our version suggests that the ``warm-up'' period is key to good accuracy.  
Overall, this suggests that choosing masks in a task-aware fashion is key to good performance in the context of LLM fine-tuning.

In summary, the experiments establish the fact that RoSA and QRoSA can indeed be competitive with the much more expensive FFT process in terms of top accuracy, while having a much lighter memory and computational footprint. This is enabled by our specific mask choice process, as well as by the efficient system support.

\paragraph{Runtime.} Performing measurements on an NVIDIA RTX A6000 GPU, we find our current implementatio  of RoSA to be approximately 1.7-2x slower than LoRA on the $80M$ parameter budget (see Appendix \ref{apdx:runtime}).
This is due to overheads in the sputnik implementation, which we plan to mitigate in future work. Furthermore, we note that fine-tuning on the down-stream tasks is usually a short process. Hence one can afford 1.7-2x slowdown compared to LoRA, considering that we are essentially able to recover FFT accuracy, and that FFT is usually either slower or not even executable in the memory-constrained setups we consider.
}

\section{Discussion}
In this paper, we took a step forward to address the problem of efficient fine-tuning of Large Language Models (LLMs). We proposed a method called Robust Adaptation (RoSA), which is inspired by the Robust PCA approach, and showed that RoSA significantly outperforms both low-rank adaptation (LoRA) \cite{hu2021lora} 
and prior sparse or hybrid approaches~\cite{sung2021training, chen2021dsee} at the same parameter budgets. Additionally, we came across the surprising observation that the best-performing RoSA can match or even outperform FFT in many settings.
To complement our contributions, we provide an efficient PyTorch implementation of our method, aiming to make RoSA an accessible tool for researchers in the field.

\section*{Acknowledgments}
The authors would like to thank Eldar Kurtic for experimental support and useful suggestions throughout the project.

\section*{Impact Statement}
This paper presents work whose goal is to advance the field of Machine Learning. There are many potential societal consequences of our work, none of which we feel must be specifically highlighted here.

\bibliographystyle{icml2024}
\bibliography{example_paper}

\begin{thebibliography}{69}
\providecommand{\natexlab}[1]{#1}
\providecommand{\url}[1]{\texttt{#1}}
\expandafter\ifx\csname urlstyle\endcsname\relax
  \providecommand{\doi}[1]{doi: #1}\else
  \providecommand{\doi}{doi: \begingroup \urlstyle{rm}\Url}\fi

\bibitem[Alex et~al.(2021)Alex, Lifland, Tunstall, Thakur, Maham, Riedel, Hine, Ashurst, Sedille, Carlier, et~al.]{alex2021raft}
Alex, N., Lifland, E., Tunstall, L., Thakur, A., Maham, P., Riedel, C.~J., Hine, E., Ashurst, C., Sedille, P., Carlier, A., et~al.
\newblock Raft: A real-world few-shot text classification benchmark.
\newblock \emph{arXiv preprint arXiv:2109.14076}, 2021.

\bibitem[Askell et~al.(2021)Askell, Bai, Chen, Drain, Ganguli, Henighan, Jones, Joseph, Mann, DasSarma, et~al.]{askell2021general}
Askell, A., Bai, Y., Chen, A., Drain, D., Ganguli, D., Henighan, T., Jones, A., Joseph, N., Mann, B., DasSarma, N., et~al.
\newblock A general language assistant as a laboratory for alignment.
\newblock \emph{arXiv preprint arXiv:2112.00861}, 2021.

\bibitem[Bai et~al.(2022)Bai, Jones, Ndousse, Askell, Chen, DasSarma, Drain, Fort, Ganguli, Henighan, et~al.]{bai2022training}
Bai, Y., Jones, A., Ndousse, K., Askell, A., Chen, A., DasSarma, N., Drain, D., Fort, S., Ganguli, D., Henighan, T., et~al.
\newblock Training a helpful and harmless assistant with reinforcement learning from human feedback.
\newblock \emph{arXiv preprint arXiv:2204.05862}, 2022.

\bibitem[Cand{\`e}s et~al.(2011)Cand{\`e}s, Li, Ma, and Wright]{candes2011robust}
Cand{\`e}s, E.~J., Li, X., Ma, Y., and Wright, J.
\newblock Robust principal component analysis?
\newblock \emph{Journal of the ACM (JACM)}, 58\penalty0 (3):\penalty0 1--37, 2011.

\bibitem[Castro et~al.(2023)Castro, Ivanov, Andrade, Ben-Nun, Fraguela, and Hoefler]{castro2023venom}
Castro, R.~L., Ivanov, A., Andrade, D., Ben-Nun, T., Fraguela, B.~B., and Hoefler, T.
\newblock Venom: A vectorized n: M format for unleashing the power of sparse tensor cores.
\newblock In \emph{Proceedings of the International Conference for High Performance Computing, Networking, Storage and Analysis}, pp.\  1--14, 2023.

\bibitem[Chen et~al.(2021)Chen, Chen, Chen, Awadallah, Wang, and Cheng]{chen2021dsee}
Chen, X., Chen, T., Chen, W., Awadallah, A.~H., Wang, Z., and Cheng, Y.
\newblock Dsee: Dually sparsity-embedded efficient tuning of pre-trained language models.
\newblock \emph{arXiv preprint arXiv:2111.00160}, 2021.

\bibitem[Cobbe et~al.(2021)Cobbe, Kosaraju, Bavarian, Chen, Jun, Kaiser, Plappert, Tworek, Hilton, Nakano, Hesse, and Schulman]{gsm8k}
Cobbe, K., Kosaraju, V., Bavarian, M., Chen, M., Jun, H., Kaiser, L., Plappert, M., Tworek, J., Hilton, J., Nakano, R., Hesse, C., and Schulman, J.
\newblock Training verifiers to solve math word problems.
\newblock \emph{arXiv preprint arXiv:2110.14168}, 2021.

\bibitem[De~La~Torre \& Black(2003)De~La~Torre and Black]{rpcainf1}
De~La~Torre, F. and Black, M.~J.
\newblock A framework for robust subspace learning.
\newblock \emph{International Journal of Computer Vision}, 54:\penalty0 117--142, 2003.

\bibitem[Dean et~al.(2012)Dean, Corrado, Monga, Chen, Devin, Mao, Ranzato, Senior, Tucker, Yang, et~al.]{bfloat16}
Dean, J., Corrado, G., Monga, R., Chen, K., Devin, M., Mao, M., Ranzato, M., Senior, A., Tucker, P., Yang, K., et~al.
\newblock Large scale distributed deep networks.
\newblock \emph{Advances in neural information processing systems}, 25, 2012.

\bibitem[Dettmers et~al.(2022)Dettmers, Lewis, Belkada, and Zettlemoyer]{dettmers2022llm}
Dettmers, T., Lewis, M., Belkada, Y., and Zettlemoyer, L.
\newblock {LLM}.int8(): 8-bit matrix multiplication for transformers at scale.
\newblock \emph{Advances in Neural Information Processing Systems 35: Annual Conference on Neural Information Processing Systems 2022, NeurIPS 2022}, 2022.

\bibitem[Dettmers et~al.(2023{\natexlab{a}})Dettmers, Pagnoni, Holtzman, and Zettlemoyer]{dettmers2023qlora}
Dettmers, T., Pagnoni, A., Holtzman, A., and Zettlemoyer, L.
\newblock Qlora: Efficient finetuning of quantized llms.
\newblock \emph{arXiv preprint arXiv:2305.14314}, 2023{\natexlab{a}}.

\bibitem[Dettmers et~al.(2023{\natexlab{b}})Dettmers, Svirschevski, Egiazarian, Kuznedelev, Frantar, Ashkboos, Borzunov, Hoefler, and Alistarh]{dettmers2023spqr}
Dettmers, T., Svirschevski, R., Egiazarian, V., Kuznedelev, D., Frantar, E., Ashkboos, S., Borzunov, A., Hoefler, T., and Alistarh, D.
\newblock Spqr: A sparse-quantized representation for near-lossless llm weight compression.
\newblock \emph{arXiv preprint arXiv:2306.03078}, 2023{\natexlab{b}}.

\bibitem[Devlin et~al.(2018)Devlin, Chang, Lee, and Toutanova]{devlin2018bert}
Devlin, J., Chang, M.-W., Lee, K., and Toutanova, K.
\newblock Bert: Pre-training of deep bidirectional transformers for language understanding.
\newblock \emph{arXiv preprint arXiv:1810.04805}, 2018.

\bibitem[Edalati et~al.(2022)Edalati, Tahaei, Kobyzev, Nia, Clark, and Rezagholizadeh]{krona}
Edalati, A., Tahaei, M., Kobyzev, I., Nia, V.~P., Clark, J.~J., and Rezagholizadeh, M.
\newblock Krona: Parameter efficient tuning with kronecker adapter.
\newblock \emph{arXiv preprint arXiv:2212.10650}, 2022.

\bibitem[Evci et~al.(2020)Evci, Gale, Menick, Castro, and Elsen]{evci2020rigging}
Evci, U., Gale, T., Menick, J., Castro, P.~S., and Elsen, E.
\newblock Rigging the lottery: Making all tickets winners.
\newblock In \emph{International Conference on Machine Learning}, pp.\  2943--2952. PMLR, 2020.

\bibitem[Fischler \& Bolles(1981)Fischler and Bolles]{rpcaransam}
Fischler, M.~A. and Bolles, R.~C.
\newblock Random sample consensus: a paradigm for model fitting with applications to image analysis and automated cartography.
\newblock \emph{Communications of the ACM}, 24\penalty0 (6):\penalty0 381--395, 1981.

\bibitem[Frantar \& Alistarh(2022)Frantar and Alistarh]{frantar2022optimal}
Frantar, E. and Alistarh, D.
\newblock Optimal brain compression: A framework for accurate post-training quantization and pruning.
\newblock \emph{Advances in Neural Information Processing Systems}, 35:\penalty0 4475--4488, 2022.

\bibitem[Gale et~al.(2019)Gale, Elsen, and Hooker]{gale2019state}
Gale, T., Elsen, E., and Hooker, S.
\newblock The state of sparsity in deep neural networks.
\newblock \emph{arXiv preprint arXiv:1902.09574}, 2019.

\bibitem[Gale et~al.(2020)Gale, Zaharia, Young, and Elsen]{sputnik}
Gale, T., Zaharia, M., Young, C., and Elsen, E.
\newblock Sparse {GPU} kernels for deep learning.
\newblock In \emph{Proceedings of the International Conference for High Performance Computing, Networking, Storage and Analysis, {SC} 2020}, 2020.

\bibitem[Gnanadesikan \& Kettenring(1972)Gnanadesikan and Kettenring]{rpcamultrim}
Gnanadesikan, R. and Kettenring, J.~R.
\newblock Robust estimates, residuals, and outlier detection with multiresponse data.
\newblock \emph{Biometrics}, pp.\  81--124, 1972.

\bibitem[Gray et~al.(2017)Gray, Radford, and Kingma]{openai}
Gray, S., Radford, A., and Kingma, D.~P.
\newblock Gpu kernels for block-sparse weights.
\newblock \emph{arXiv preprint arXiv:1711.09224}, 3\penalty0 (2):\penalty0 2, 2017.

\bibitem[He et~al.(2022)He, Zhou, Ma, Berg-Kirkpatrick, and Neubig]{he2022towards}
He, J., Zhou, C., Ma, X., Berg-Kirkpatrick, T., and Neubig, G.
\newblock Towards a unified view of parameter-efficient transfer learning.
\newblock In \emph{Proceedings of the 10th International Conference on Learning Representations (ICLR-2022)}, 2022.

\bibitem[Hendrycks et~al.(2020)Hendrycks, Burns, Basart, Zou, Mazeika, Song, and Steinhardt]{hendrycksmeasuring}
Hendrycks, D., Burns, C., Basart, S., Zou, A., Mazeika, M., Song, D., and Steinhardt, J.
\newblock Measuring massive multitask language understanding.
\newblock In \emph{International Conference on Learning Representations}, 2020.

\bibitem[Hoefler et~al.(2021)Hoefler, Alistarh, Ben-Nun, Dryden, and Peste]{hoefler2021sparsity}
Hoefler, T., Alistarh, D., Ben-Nun, T., Dryden, N., and Peste, A.
\newblock Sparsity in deep learning: Pruning and growth for efficient inference and training in neural networks.
\newblock \emph{The Journal of Machine Learning Research}, 22\penalty0 (1):\penalty0 10882--11005, 2021.

\bibitem[Hu et~al.(2021)Hu, Shen, Wallis, Allen-Zhu, Li, Wang, Wang, and Chen]{hu2021lora}
Hu, E.~J., Shen, Y., Wallis, P., Allen-Zhu, Z., Li, Y., Wang, S., Wang, L., and Chen, W.
\newblock Lora: Low-rank adaptation of large language models.
\newblock \emph{arXiv preprint arXiv:2106.09685}, 2021.

\bibitem[Hubara et~al.(2021)Hubara, Chmiel, Island, Banner, Naor, and Soudry]{hubara2021accelerated}
Hubara, I., Chmiel, B., Island, M., Banner, R., Naor, J., and Soudry, D.
\newblock Accelerated sparse neural training: A provable and efficient method to find n: m transposable masks.
\newblock \emph{Advances in neural information processing systems}, 34:\penalty0 21099--21111, 2021.

\bibitem[Huber(2004)]{rpcainf2}
Huber, P.~J.
\newblock \emph{Robust statistics}, volume 523.
\newblock John Wiley \& Sons, 2004.

\bibitem[Hyeon-Woo et~al.(2021)Hyeon-Woo, Ye-Bin, and Oh]{peft7}
Hyeon-Woo, N., Ye-Bin, M., and Oh, T.-H.
\newblock Fedpara: Low-rank hadamard product for communication-efficient federated learning.
\newblock \emph{arXiv preprint arXiv:2108.06098}, 2021.

\bibitem[Ivanov et~al.(2022)Ivanov, Dryden, and Hoefler]{ivanov2022sten}
Ivanov, A., Dryden, N., and Hoefler, T.
\newblock Sten: An interface for efficient sparsity in pytorch.
\newblock 2022.

\bibitem[Jiang et~al.(2022)Jiang, Hu, and Song]{jiang2022exposing}
Jiang, P., Hu, L., and Song, S.
\newblock Exposing and exploiting fine-grained block structures for fast and accurate sparse training.
\newblock \emph{Advances in Neural Information Processing Systems}, 35:\penalty0 38345--38357, 2022.

\bibitem[Juraska et~al.(2019)Juraska, Bowden, and Walker]{viggo}
Juraska, J., Bowden, K., and Walker, M.
\newblock {V}i{GGO}: A video game corpus for data-to-text generation in open-domain conversation.
\newblock In \emph{Proceedings of the 12th International Conference on Natural Language Generation}, pp.\  164--172, Tokyo, Japan, October{--}November 2019. Association for Computational Linguistics.
\newblock \doi{10.18653/v1/W19-8623}.
\newblock URL \url{https://aclanthology.org/W19-8623}.

\bibitem[Ke \& Kanade(2005)Ke and Kanade]{rpcaaltmin}
Ke, Q. and Kanade, T.
\newblock Robust l/sub 1/norm factorization in the presence of outliers and missing data by alternative convex programming.
\newblock In \emph{2005 IEEE Computer Society Conference on Computer Vision and Pattern Recognition (CVPR'05)}, volume~1, pp.\  739--746. IEEE, 2005.

\bibitem[Kurtic et~al.(2022)Kurtic, Campos, Nguyen, Frantar, Kurtz, Fineran, Goin, and Alistarh]{kurtic2022optimal}
Kurtic, E., Campos, D., Nguyen, T., Frantar, E., Kurtz, M., Fineran, B., Goin, M., and Alistarh, D.
\newblock The optimal bert surgeon: Scalable and accurate second-order pruning for large language models.
\newblock \emph{arXiv preprint arXiv:2203.07259}, 2022.

\bibitem[Kurtic et~al.(2023)Kurtic, Kuznedelev, Frantar, Goin, and Alistarh]{kurtic2023sparse}
Kurtic, E., Kuznedelev, D., Frantar, E., Goin, M., and Alistarh, D.
\newblock Sparse finetuning for inference acceleration of large language models.
\newblock \emph{arXiv preprint arXiv:2310.06927}, 2023.

\bibitem[Lee et~al.(2023)Lee, Hunter, and Ruiz]{lee2023platypus}
Lee, A.~N., Hunter, C.~J., and Ruiz, N.
\newblock Platypus: Quick, cheap, and powerful refinement of llms.
\newblock \emph{arXiv preprint arXiv:2308.07317}, 2023.

\bibitem[Lester et~al.(2021)Lester, Al-Rfou, and Constant]{lester2021power}
Lester, B., Al-Rfou, R., and Constant, N.
\newblock The power of scale for parameter-efficient prompt tuning.
\newblock \emph{arXiv preprint arXiv:2104.08691}, 2021.

\bibitem[Li et~al.(2022)Li, Osawa, and Hoefler]{magicube}
Li, S., Osawa, K., and Hoefler, T.
\newblock Efficient quantized sparse matrix operations on tensor cores.
\newblock In \emph{SC22: International Conference for High Performance Computing, Networking, Storage and Analysis}, pp.\  1--15. IEEE, 2022.

\bibitem[Li \& Liang(2021)Li and Liang]{li2021prefix}
Li, X.~L. and Liang, P.
\newblock Prefix-tuning: Optimizing continuous prompts for generation.
\newblock \emph{arXiv preprint arXiv:2101.00190}, 2021.

\bibitem[Li et~al.(2023)Li, Yu, Liang, He, Karampatziakis, Chen, and Zhao]{peft9}
Li, Y., Yu, Y., Liang, C., He, P., Karampatziakis, N., Chen, W., and Zhao, T.
\newblock Loftq: Lora-fine-tuning-aware quantization for large language models.
\newblock \emph{arXiv preprint arXiv:2310.08659}, 2023.

\bibitem[Liu et~al.(2022)Liu, Tam, Muqeeth, Mohta, Huang, Bansal, and Raffel]{liu2022few}
Liu, H., Tam, D., Muqeeth, M., Mohta, J., Huang, T., Bansal, M., and Raffel, C.~A.
\newblock Few-shot parameter-efficient fine-tuning is better and cheaper than in-context learning.
\newblock \emph{Advances in Neural Information Processing Systems}, 35:\penalty0 1950--1965, 2022.

\bibitem[Liu et~al.(2021)Liu, Ji, Fu, Tam, Du, Yang, and Tang]{peft2}
Liu, X., Ji, K., Fu, Y., Tam, W.~L., Du, Z., Yang, Z., and Tang, J.
\newblock P-tuning v2: Prompt tuning can be comparable to fine-tuning universally across scales and tasks.
\newblock \emph{arXiv preprint arXiv:2110.07602}, 2021.

\bibitem[Liu et~al.(2023)Liu, Zheng, Du, Ding, Qian, Yang, and Tang]{peft3}
Liu, X., Zheng, Y., Du, Z., Ding, M., Qian, Y., Yang, Z., and Tang, J.
\newblock Gpt understands, too.
\newblock \emph{AI Open}, 2023.

\bibitem[Loshchilov \& Hutter(2017)Loshchilov and Hutter]{loshchilov2017decoupled}
Loshchilov, I. and Hutter, F.
\newblock Decoupled weight decay regularization.
\newblock \emph{arXiv preprint arXiv:1711.05101}, 2017.

\bibitem[Mangrulkar et~al.(2022)Mangrulkar, Gugger, Debut, Belkada, Paul, and Bossan]{peftlibrary}
Mangrulkar, S., Gugger, S., Debut, L., Belkada, Y., Paul, S., and Bossan, B.
\newblock Peft: State-of-the-art parameter-efficient fine-tuning methods.
\newblock \url{https://github.com/huggingface/peft}, 2022.

\bibitem[Min et~al.(2021)Min, Lewis, Zettlemoyer, and Hajishirzi]{min2021metaicl}
Min, S., Lewis, M., Zettlemoyer, L., and Hajishirzi, H.
\newblock Metaicl: Learning to learn in context.
\newblock \emph{arXiv preprint arXiv:2110.15943}, 2021.

\bibitem[MosaicML(2023{\natexlab{a}})]{llm-foundry}
MosaicML.
\newblock {LLM Foundry}, 2023{\natexlab{a}}.
\newblock URL \url{https://github.com/mosaicml/llm-foundry}.

\bibitem[MosaicML(2023{\natexlab{b}})]{mpt}
MosaicML.
\newblock Introducing mpt-7b: A new standard for open-source, commercially usable llms, 2023{\natexlab{b}}.
\newblock URL \url{www.mosaicml.com/blog/mpt-7b}.
\newblock Accessed: 2023-12-22.

\bibitem[Niederfahrenhorst et~al.(2023)Niederfahrenhorst, Hakhamaneshi, and Ahmad]{Anyscale}
Niederfahrenhorst, A., Hakhamaneshi, K., and Ahmad, R.
\newblock {Fine-Tuning LLMs: LoRA or Full-Parameter?}, 2023.
\newblock URL \url{https://www.anyscale.com/blog/fine-tuning-llms-lora-or-full-parameter-an-in-depth-analysis-with-llama-2}.

\bibitem[Nikdan et~al.(2023)Nikdan, Pegolotti, Iofinova, Kurtic, and Alistarh]{nikdan2023sparseprop}
Nikdan, M., Pegolotti, T., Iofinova, E., Kurtic, E., and Alistarh, D.
\newblock Sparseprop: Efficient sparse backpropagation for faster training of neural networks at the edge.
\newblock In \emph{International Conference on Machine Learning}, pp.\  26215--26227. PMLR, 2023.

\bibitem[Ouyang et~al.(2022)Ouyang, Wu, Jiang, Almeida, Wainwright, Mishkin, Zhang, Agarwal, Slama, Ray, et~al.]{ouyang2022training}
Ouyang, L., Wu, J., Jiang, X., Almeida, D., Wainwright, C., Mishkin, P., Zhang, C., Agarwal, S., Slama, K., Ray, A., et~al.
\newblock Training language models to follow instructions with human feedback.
\newblock \emph{Advances in Neural Information Processing Systems}, 35:\penalty0 27730--27744, 2022.

\bibitem[Paszke et~al.(2019)Paszke, Gross, Massa, Lerer, Bradbury, Chanan, Killeen, Lin, Gimelshein, Antiga, et~al.]{paszke2019pytorch}
Paszke, A., Gross, S., Massa, F., Lerer, A., Bradbury, J., Chanan, G., Killeen, T., Lin, Z., Gimelshein, N., Antiga, L., et~al.
\newblock Pytorch: An imperative style, high-performance deep learning library.
\newblock In \emph{Advances in Neural Information Processing Systems}, 2019.

\bibitem[Peste et~al.(2021)Peste, Iofinova, Vladu, and Alistarh]{peste2021ac}
Peste, A., Iofinova, E., Vladu, A., and Alistarh, D.
\newblock Ac/dc: Alternating compressed/decompressed training of deep neural networks.
\newblock \emph{Advances in neural information processing systems}, 34:\penalty0 8557--8570, 2021.

\bibitem[Qiu et~al.(2023)Qiu, Liu, Feng, Xue, Feng, Liu, Zhang, Weller, and Sch{\"o}lkopf]{peft10}
Qiu, Z., Liu, W., Feng, H., Xue, Y., Feng, Y., Liu, Z., Zhang, D., Weller, A., and Sch{\"o}lkopf, B.
\newblock Controlling text-to-image diffusion by orthogonal finetuning.
\newblock \emph{arXiv preprint arXiv:2306.07280}, 2023.

\bibitem[Sanh et~al.(2020)Sanh, Wolf, and Rush]{sanh2020movement}
Sanh, V., Wolf, T., and Rush, A.
\newblock Movement pruning: Adaptive sparsity by fine-tuning.
\newblock \emph{Advances in Neural Information Processing Systems}, 33:\penalty0 20378--20389, 2020.

\bibitem[Sanh et~al.(2021)Sanh, Webson, Raffel, Bach, Sutawika, Alyafeai, Chaffin, Stiegler, Scao, Raja, et~al.]{sanh2021multitask}
Sanh, V., Webson, A., Raffel, C., Bach, S.~H., Sutawika, L., Alyafeai, Z., Chaffin, A., Stiegler, A., Scao, T.~L., Raja, A., et~al.
\newblock Multitask prompted training enables zero-shot task generalization.
\newblock \emph{arXiv preprint arXiv:2110.08207}, 2021.

\bibitem[Singh \& Alistarh(2020)Singh and Alistarh]{singh2020woodfisher}
Singh, S.~P. and Alistarh, D.
\newblock Woodfisher: Efficient second-order approximation for neural network compression.
\newblock \emph{Advances in Neural Information Processing Systems}, 33:\penalty0 18098--18109, 2020.

\bibitem[Sung et~al.(2021)Sung, Nair, and Raffel]{sung2021training}
Sung, Y.-L., Nair, V., and Raffel, C.~A.
\newblock Training neural networks with fixed sparse masks.
\newblock \emph{Advances in Neural Information Processing Systems}, 34:\penalty0 24193--24205, 2021.

\bibitem[Taori et~al.(2023)Taori, Gulrajani, Zhang, Dubois, Li, Guestrin, Liang, and Hashimoto]{alpaca}
Taori, R., Gulrajani, I., Zhang, T., Dubois, Y., Li, X., Guestrin, C., Liang, P., and Hashimoto, T.~B.
\newblock Stanford alpaca: An instruction-following llama model.
\newblock \url{https://github.com/tatsu-lab/stanford_alpaca}, 2023.

\bibitem[Touvron et~al.(2023{\natexlab{a}})Touvron, Lavril, Izacard, Martinet, Lachaux, Lacroix, Rozi{\`e}re, Goyal, Hambro, Azhar, et~al.]{touvron2023llama}
Touvron, H., Lavril, T., Izacard, G., Martinet, X., Lachaux, M.-A., Lacroix, T., Rozi{\`e}re, B., Goyal, N., Hambro, E., Azhar, F., et~al.
\newblock Llama: Open and efficient foundation language models.
\newblock \emph{arXiv preprint arXiv:2302.13971}, 2023{\natexlab{a}}.

\bibitem[Touvron et~al.(2023{\natexlab{b}})Touvron, Martin, Stone, Albert, Almahairi, Babaei, Bashlykov, Batra, Bhargava, Bhosale, et~al.]{touvron2023llama2}
Touvron, H., Martin, L., Stone, K., Albert, P., Almahairi, A., Babaei, Y., Bashlykov, N., Batra, S., Bhargava, P., Bhosale, S., et~al.
\newblock Llama 2: Open foundation and fine-tuned chat models.
\newblock \emph{arXiv preprint arXiv:2307.09288}, 2023{\natexlab{b}}.

\bibitem[Wang et~al.(2022{\natexlab{a}})Wang, Kordi, Mishra, Liu, Smith, Khashabi, and Hajishirzi]{wang2022self}
Wang, Y., Kordi, Y., Mishra, S., Liu, A., Smith, N.~A., Khashabi, D., and Hajishirzi, H.
\newblock Self-instruct: Aligning language model with self generated instructions.
\newblock \emph{arXiv preprint arXiv:2212.10560}, 2022{\natexlab{a}}.

\bibitem[Wang et~al.(2022{\natexlab{b}})Wang, Mishra, Alipoormolabashi, Kordi, Mirzaei, Naik, Ashok, Dhanasekaran, Arunkumar, Stap, et~al.]{wang2022super}
Wang, Y., Mishra, S., Alipoormolabashi, P., Kordi, Y., Mirzaei, A., Naik, A., Ashok, A., Dhanasekaran, A.~S., Arunkumar, A., Stap, D., et~al.
\newblock Super-naturalinstructions: Generalization via declarative instructions on 1600+ nlp tasks.
\newblock In \emph{Proceedings of the 2022 Conference on Empirical Methods in Natural Language Processing}, pp.\  5085--5109, 2022{\natexlab{b}}.

\bibitem[Wei et~al.(2021)Wei, Bosma, Zhao, Guu, Yu, Lester, Du, Dai, and Le]{wei2021finetuned}
Wei, J., Bosma, M., Zhao, V.~Y., Guu, K., Yu, A.~W., Lester, B., Du, N., Dai, A.~M., and Le, Q.~V.
\newblock Finetuned language models are zero-shot learners.
\newblock \emph{arXiv preprint arXiv:2109.01652}, 2021.

\bibitem[Wright et~al.(2009)Wright, Ganesh, Rao, Peng, and Ma]{wright2009robust}
Wright, J., Ganesh, A., Rao, S., Peng, Y., and Ma, Y.
\newblock Robust principal component analysis: Exact recovery of corrupted low-rank matrices via convex optimization.
\newblock \emph{Advances in neural information processing systems}, 22, 2009.

\bibitem[Yu et~al.(2018)Yu, Zhang, Yang, Yasunaga, Wang, Li, Ma, Li, Yao, Roman, et~al.]{yu2018spider}
Yu, T., Zhang, R., Yang, K., Yasunaga, M., Wang, D., Li, Z., Ma, J., Li, I., Yao, Q., Roman, S., et~al.
\newblock Spider: A large-scale human-labeled dataset for complex and cross-domain semantic parsing and text-to-sql task.
\newblock \emph{arXiv preprint arXiv:1809.08887}, 2018.

\bibitem[Zhang et~al.(2023)Zhang, Chen, Bukharin, He, Cheng, Chen, and Zhao]{adalora}
Zhang, Q., Chen, M., Bukharin, A., He, P., Cheng, Y., Chen, W., and Zhao, T.
\newblock Adaptive budget allocation for parameter-efficient fine-tuning.
\newblock \emph{arXiv preprint arXiv:2303.10512}, 2023.

\bibitem[Zhang et~al.(2022)Zhang, Roller, Goyal, Artetxe, Chen, Chen, Dewan, Diab, Li, Lin, et~al.]{opt}
Zhang, S., Roller, S., Goyal, N., Artetxe, M., Chen, M., Chen, S., Dewan, C., Diab, M., Li, X., Lin, X.~V., et~al.
\newblock Opt: Open pre-trained transformer language models.
\newblock \emph{arXiv preprint arXiv:2205.01068}, 2022.

\bibitem[Zhong et~al.(2017)Zhong, Xiong, and Socher]{Seq2SQL}
Zhong, V., Xiong, C., and Socher, R.
\newblock Seq2sql: Generating structured queries from natural language using reinforcement learning.
\newblock \emph{CoRR}, abs/1709.00103, 2017.

\bibitem[Zhou \& Tao(2013)Zhou and Tao]{grebsmo-zhou13b}
Zhou, T. and Tao, D.
\newblock Greedy bilateral sketch, completion \& smoothing.
\newblock In \emph{Proceedings of the Sixteenth International Conference on Artificial Intelligence and Statistics}, volume~31 of \emph{Proceedings of Machine Learning Research}, pp.\  650--658. PMLR, 2013.

\end{thebibliography}

\newpage
\appendix 
\onecolumn

\section{System Details}
\label{apdx:system}
We integrated RoSA into a fork of the standard \texttt{peft} library \cite{peftlibrary}, and performed all the experiments using the the \texttt{llm-foundry} codebase \cite{llm-foundry}. Next, we will elaborate on the efficient implementation of RoSA.

\begin{figure}[t]
  \centering
  \includegraphics[width=1\columnwidth]{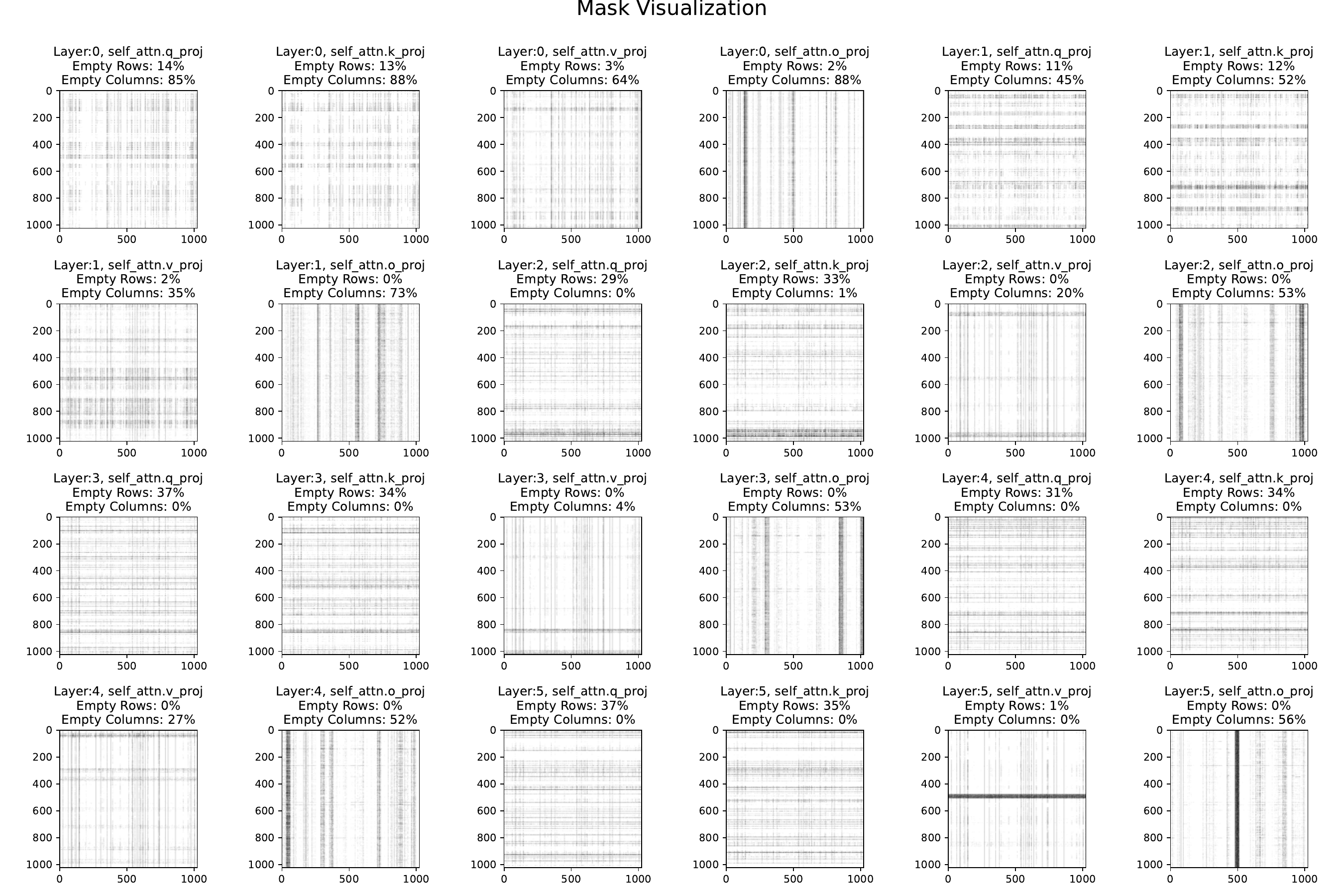}
  \caption{Here we see a visualization of a subset of masks taken from LLaMa2-7B Model trained on GSM8k ($r=16, d=0.6\%$). We can see that most masks visualized here have either a significant number of empty rows or columns. For the purposes of visualization, each mask is max-pooled with a kernel size and stride of 4. }
  \label{fig:masks}
\end{figure}

\paragraph{\textbf{Mask Structure.}} As noted in Section \ref{sec:sys}, our findings show that a significant number of either mask rows or columns are completely empty. Figure \ref{fig:masks} shows a visualization of this phenomenon, and Table \ref{tab:empty_masks} outlines the empty rows across a wider range of models over a subset of our models. It shows, for each model, the mean of the maximum percentage of empty rows or columns. Finally, we report that a mean of $ 46.74\% $ (rounded to two decimals) of the maximum between the percentage of empty rows or columns is present across all of our trained models. The prevalence of empty rows and columns emphasizes the motivation to use a kernel that does not launch threads for outputs where no work is needed.

\begin{table*}[h]
\centering
\caption{This table shows the row and column statistics for a subset of the models with across a wide range of datasets and densities. Note that the masks depend on the learning rate because they were generated after a LoRA warmup period.}
\scalebox{0.8}{
\begin{tabular}{cccc}
LLaMA 7B & Maximal Empty Row & Maximal Empty Column & Mean Maximal Empty Row or Column \\
\toprule
\texttt{GSM8K} & & & \\
 $ d = 0.0015, r= 12, lr = 0.0002 $ & $ 98.18\% $ & $ 98.97\% $ & $ 73.49\% $ \\
 $ d = 0.003, r= 8, lr = 0.0002 $ & $ 97.85\% $ & $ 96.72\% $ & $ 58.30\% $ \\
 $ d = 0.006, r= 48, lr = 0.0002 $ & $ 97.5\% $ & $ 94.03\% $ & $ 40.94\% $ \\
 $ d = 0.012, r= 32, lr = 0.0002 $ & $ 96.46\% $ & $ 85.01\% $ & $ 27.12\% $ \\
 $ d = 0.018, r= 16, lr = 0.0004 $ & $ 94.79\% $ & $ 79.94\% $ & $ 19.60\% $ \\
\toprule
\texttt{SQL} & & & \\
$ d = 0.0015, r= 12, lr = 0.0004 $ & $ 99.14\% $ & $ 97.92\% $ & $ 79.34\% $ \\
$ d = 0.003, r= 8, lr = 0.0004 $ & $ 98.61\% $ & $ 96.72\% $ & $ 65.94\% $ \\
$ d = 0.0045, r= 4, lr = 0.0004 $ & $ 97.96\% $ & $ 95.70\% $ & $ 56.36\% $ \\
$ d = 0.006, r= 48, lr = 0.0004 $ & $ 96.56\% $ & $ 94.10\% $ & $ 48.84\% $ \\
$ d = 0.009, r= 8, lr = 0.0001 $ & $ 95.32\% $ & $ 87.28\% $ & $ 41.25\% $ \\
$ d = 0.012, r= 32, lr = 0.0004 $ & $ 91.13\% $ & $ 85.15\% $ & $ 34.46\% $ \\
$ d = 0.018, r= 16, lr = 0.0004 $ & $ 86.87\% $ & $ 80.06\% $ & $ 29.74\% $ \\
\toprule
\texttt{ViGGO} & & & \\
$ d = 0.0015, r= 12, lr = 0.0002 $ & $ 99.53\% $ & $ 98.90\% $ & $ 75.29\% $ \\
$ d = 0.003, r= 8, lr = 0.0002 $ & $ 99.04\% $ & $ 97.50\% $ & $ 61.68\% $ \\
$ d = 0.0045, r= 4, lr = 0.0002 $ & $ 96.19\% $ & $ 91.43\% $ & $ 55.22\% $ \\
$ d = 0.006, r= 48, lr = 0.0002 $ & $ 91.38\% $ & $ 90.91\% $ & $ 46.14\% $ \\
$ d = 0.009, r= 8, lr = 0.0002 $ & $ 95.27\% $ & $ 92.19\% $ & $ 37.95\% $ \\
$ d = 0.012, r= 32, lr = 0.0002 $ & $ 94.28\% $ & $ 87.32\% $ & $ 30.88\% $ \\
$ d = 0.018, r= 16, lr = 0.0002 $ & $ 92.11\% $ & $ 82.88\% $ & $ 24.11\% $ \\
\toprule
\end{tabular}
} \label{tab:empty_masks}
\end{table*}

\subsection{SDDMM Kernel}
Our SDDMM kernel is based on the \texttt{sputnik} kernel \cite{sputnik}. Their original SDDMM implementation was extended in two ways. First, the original SDDMM kernel, as noted in the referenced publication, launches the maximum number of threads over the entire output matrix and then simply terminates those threads that have no work to do. In order to accommodate the fact that a significant portion of either the rows or columns of each individual mask is empty, we limit the number of threads launched to the number of rows and columns that have a non-zero value. At first glance, this seems to contradict the original paper's claim that the extra threads don't induce significant overhead. However, the original publication did not focus on benchmarking the low sparsity and structures present in this paper. Furthermore, as row sorting according to the number of non-zero values is part of the original implementation's pipeline, the additional necessary kernel launch information can be calculated without significant overhead. Second, the SDDMM implementation was extended to support 16-bit indices. 

\begin{figure}[t]
  \centering
  \includegraphics[width=0.9\columnwidth]{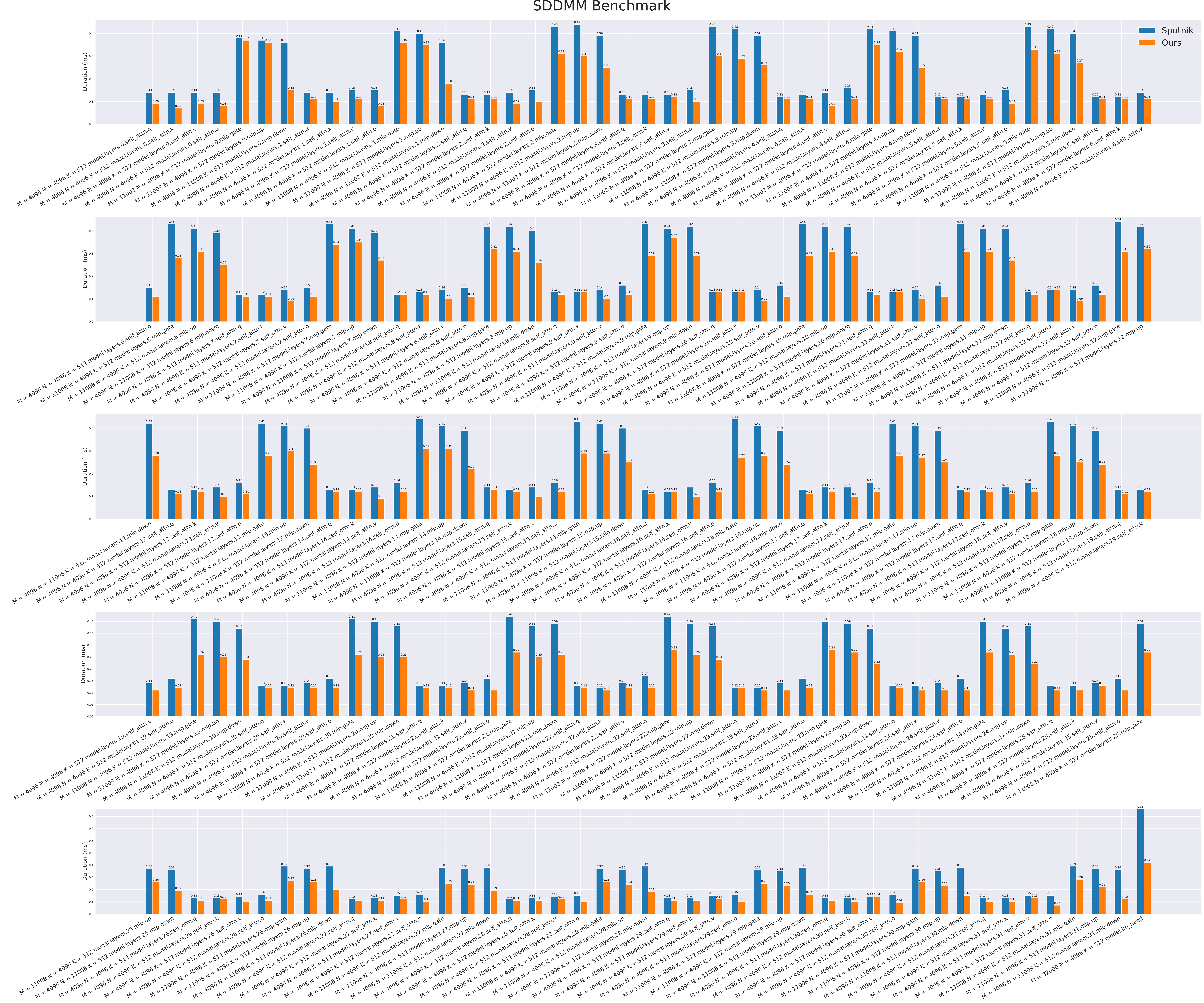}
  \caption{This figure shows the result of benchmarking SDDMM kernels with masks extracted from LLaMA2-7B $d=0.6\%$ and $ r=16$.  Compared to \texttt{sputnik} we achieve a geometric mean speedup of 1.36x and a peak speedup of 3x.}
  \label{fig:sddmm_perf}
\end{figure}

We present the benchmark results of these two changes in Figure \ref{fig:sddmm_perf}. We extract masks from  LLaMA2-7B $d=0.6\%$ and $ r=16$. For each mask $ \bM$ and construct two randomly generated \texttt{float32} matrices $ \bA $ and $ \bB$ with dimensions $ (M, K) $ and $ (N, K) $ and compute the SDDMM. We have a fixed $ K = 512 $ in this synthetic benchmark. The durations are rounded to two decimal places.

\subsection{CSR-ADD Kernel}

A CUDA kernel calculating the $\bA = \bA + \bB$ operation where $ \bA $ is dense and $ \bB $ is sparse (stored in the CSR format), was implemented with support for \texttt{float32}, \texttt{float16} and \texttt{bfloat16} input data types. It distributes thread blocks over rows of $ \bB $ with each warp, then goes over the nonzero values and adds them to the dense matrix.

\subsection{Other Details}

\paragraph{RoSA Pseudocode.}
We include a straight-forward pseudocode that describes our adaptation method (Algorithm \ref{alg:rosa}).

\begin{algorithm}[tb]
 \footnotesize
   \caption{Robust Adaptation (\methodname)}
   \label{alg:rosa}
\begin{algorithmic}
    \REQUIRE $\f{W}, \wbar \gets$ the fully connected weights and the rest of the LLM parameters, respectively
    \REQUIRE $\f{D} \gets$ the downstream dataset
    \REQUIRE $\f{L}(.) \gets$ the loss function
    \REQUIRE $r \gets$ LoRA rank
    \REQUIRE $d \gets$ SpA density
    \REQUIRE $m \gets$ number of samples to use for mask generation

    \STATE $\texttt{[m random samples for mask generation]}$
    \STATE $\f{D}_\f{M} \gets \textit{random-subset}(\f{D},m)$
    \STATE $\texttt{[run Algorithm \ref{alg:mask} to generate the masks]}$
    \STATE $\f{M} \gets \textit{generate-masks}(\f{W}, \wbar, \f{D}_\f{M}, \f{L}, d)$
    \STATE $k \gets \textit{length}(\f{W})$ 
    \FOR {$i \in \{1, 2, ..., k\}$}
        \STATE $m_i, n_i \gets \textit{shape}(\bW_i)$
        \STATE $\texttt{[init LoRA (\cite{hu2021lora})]}$
        \STATE $\bD_i^L \gets \textit{initialize-lora-params}(m_i, n_i, r)$
        \STATE $\texttt{[init SpA with zero]}$
        \STATE $\bD_i^S \gets \textit{initialize-spa-params}(\f{M}_i)$
    \ENDFOR
    \STATE $\Delta^L \gets \{\bD_1^L, \bD_2^L, ..., \bD_k^L\}$
    \STATE $\Delta^S \gets \{\bD_1^S, \bD_2^S, ..., \bD_k^S\}$
    \STATE $\texttt{[train the adapters]}$
    \STATE $\Delta^L_*, \Delta^S_* \gets \underset{\Delta^L, \Delta^S}{\argmin} ~ \f{L}(\f{D}; \f{W} + \Delta^L + \Delta^S, \wbar)$
    \STATE \textbf{return} $\Delta^L_*, \Delta^S_*$
\end{algorithmic}
\end{algorithm}

\paragraph{\textbf{Gradient Collection for QRoSA.}}
Since automatic differentiation is not supported for quantized tensors in PyTorch, in the QRoSA experiments, we manually multiply the output gradients and inputs during training to calculate the weight gradients required for mask collection.

{
\section{Runtime}
\label{apdx:runtime}
In Table \ref{tab:runtime} we compare the runtime of RoSA and LoRA on an NVIDIA RTX A6000. We observe a slow-down relative to LoRA of around 2x. 
This is because of overheads due to sparsity, but also because the sparse operators we use work with FP32 precision, which is slower than LoRA operations, which employ FP16. 
}

\begin{table}[h]
\centering
\caption{\label{tab:runtime} {Runtime comparison between (Q)LoRA and (Q)RoSA in the $80$M parameter budget. The measurements are done using an NVIDIA RTX A6000 GPU.}}
\begin{tabular}{cc}
\toprule
\textbf{Method}    & \textbf{batch/second} \\ \hline
\texttt{LoRA} $r=32$ & $0.1149$ \\ 
\texttt{RoSA} $r=24,d=0.3\%$ & $0.0602$ \\ 
\texttt{RoSA} $r=16,d=0.6\%$ & $0.0595$ \\ 
\texttt{RoSA} $r=8,d=0.9\%$ & $0.0575$ \\ 
\texttt{SpA} $d=1.2\%$ & $0.0622$ \\  \hline
\texttt{QLoRA} $r=32$ & $0.0911$ \\ 
\texttt{QRoSA} $r=24,d=0.3\%$ & $0.0531$ \\ 
\texttt{QRoSA} $r=16,d=0.6\%$ & $0.0521$ \\ 
\texttt{QRoSA} $r=8,d=0.9\%$ & $0.0515$ \\ 
\texttt{QSpA} $d=1.2\%$ & $0.0546$ \\ \toprule
\end{tabular}
\end{table}

\section{{Comparison with IA3}}
In this section, we compare our proposed method, RoSA, with IA3 \cite{liu2022few}, another parameter-efficient fine-tuning technique. IA3 involves introducing scaling parameters for the activations within a neural network.

Table \ref{tab:ia3} shows that IA3 performs poorly compared to RoSA and LoRA in terms of accuracy on the three GSM8k, ViGGO, and SQL datasets. One explanation for this underperformance is that IA3 clearly underfits due to its small parameter count. Unlike RoSA and LoRA, which introduce additional parameters through low-rank and sparse adaptations, IA3's scaling parameters are insufficient to capture the complexity of the tasks, leading to suboptimal performance.

However, it is important to note that IA3 is designed to excel in few-shot learning scenarios. For example, on the RAFT dataset \cite{alex2021raft}, which is specifically curated for few-shot learning tasks, IA3 demonstrates competitive performance. This is in contrast to RoSA and LoRA, which generally require a larger dataset to achieve optimal results.

\begin{table*}[h]
\centering
\caption{{Comparison of fine-tuning LLaMA2-7B using FFT, RoSA and IA3 \cite{liu2022few}. For RoSA, we consider $40$M, $80$M, and $160$M parameter budgets and we assume the budget is distributed equally between the sparse and low-rank adapters.}}
\label{tab:ia3}
\scalebox{0.9}{
\begin{tabular}{cccccccc}
\toprule
                  &       &         \multicolumn{2}{c}{\texttt{GSM8k}}          & \multicolumn{2}{c}{\texttt{ViGGO}}          & \multicolumn{1}{c}{\texttt{SQL}}           \\ \cline{3-7}
                  & \texttt{\#Params} & \texttt{1 Epoch} & \texttt{Extended} & \texttt{1 Epoch} & \texttt{Extended} & \texttt{1 Epoch} \\ \toprule 
\texttt{FFT} & $6.7$ \texttt{B} & $\boldsymbol{32.3}$ & $\boldsymbol{38.8}$ & $\boldsymbol{82.1}$ & $\boldsymbol{95.0}$ & $\boldsymbol{89.0}$ \\
\toprule
\texttt{RoSA} $r=8,d=0.3\%$ & $40.8$ \texttt{M} & $29.2$ & $37.5$ & $94.5$ & $97.1$ & $77.6$ \\
\texttt{RoSA} $r=16,d=0.6\%$ & $81.6$ \texttt{M} & $\boldsymbol{32.2}$ & $\boldsymbol{38.6}$ & $\boldsymbol{95.2}$ & $97.1$ & $88.3$ \\
\texttt{RoSA} $r=32,d=1.2\%$ & $163.1$ \texttt{M} & $\boldsymbol{32.2}$ & $36.2$ & $93.4$ & $\boldsymbol{97.3}$ & $\boldsymbol{89.2}$ \\
\texttt{IA3} & $1.6$ \texttt{M} & $13.12$ & $16.07$ & $38.24$ & $40.06$ & $84.5$ \\
\toprule
\end{tabular}
}
\end{table*}

\newpage
\section{Singular Value Analysis on Full Fine-Tuning}
\label{apdx:fft_pca}

\begin{figure}[ht]

  \centering
  \includegraphics[width=.82\linewidth]{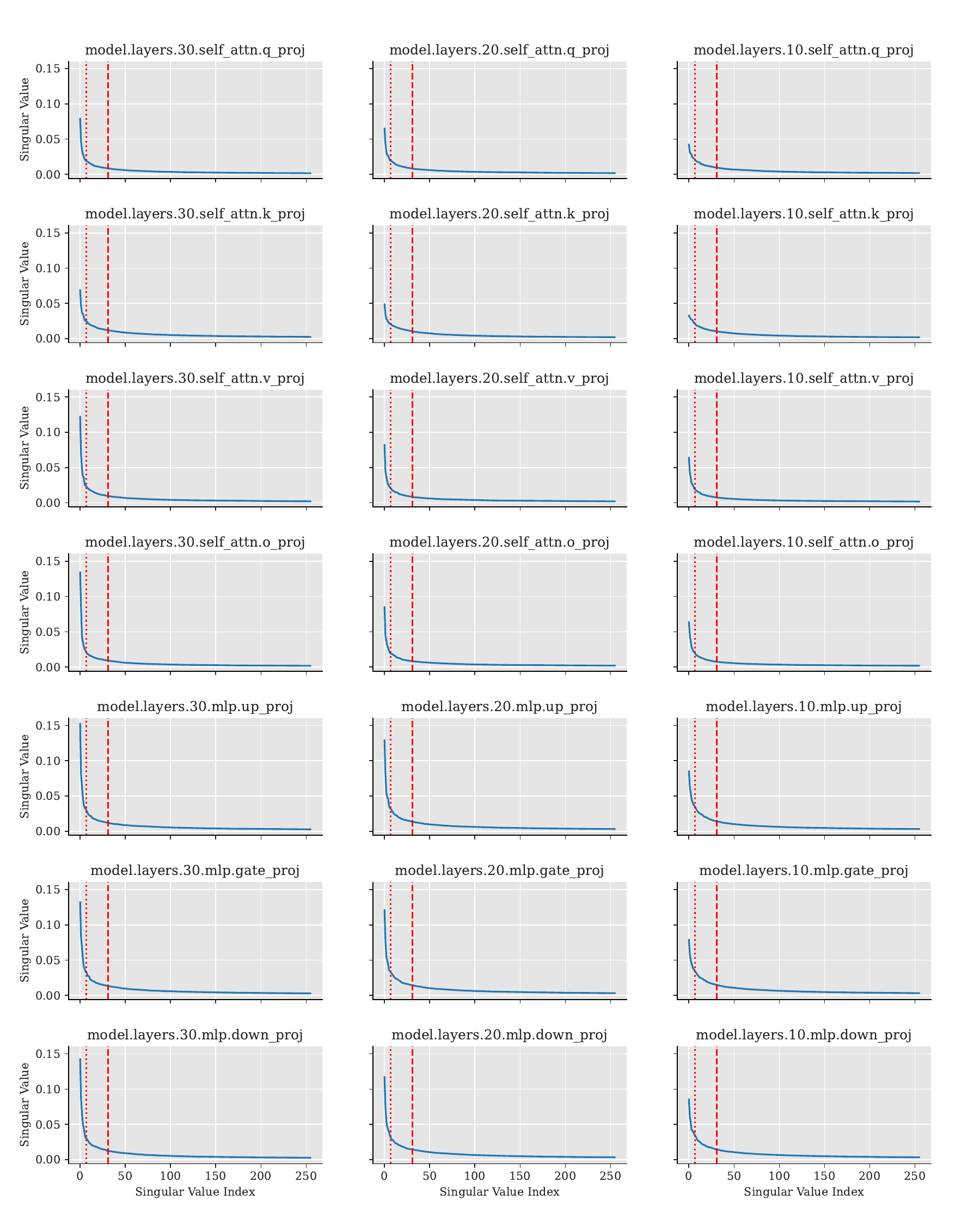}
\caption{\label{fig:diff_svd} Sorted singular values of $\Delta^*$ for various layers of a LLaMA2-7B fully fine-tuned on GSM8k. Thresholds for ranks 8 and 32 are marked with dotted and dashed lines, respectively. The top 256 singular values are selected.}
\end{figure}

We present a straightforward analysis of the singular values obtained from $\bD^*$ of the LLaMA2-7B model \cite{touvron2023llama2} fine-tuned on the GSM8k dataset. The focus is on a set of plots representing singular values from several randomly selected layers of the LLaMA2-7B model. The plots in Figure \ref{fig:diff_svd} reveal a notable pattern: a few singular values are significantly larger compared to the rest, which is relatively small yet not zero.

This pattern in the singular values suggests that the updates made during full fine-tuning of LLaMA2 exhibit a tendency towards a low-rank structure. However, they cannot be considered purely low-rank due to the presence of these small, non-zero singular values.

\section{{Instruction-tuning Results}}
In this section, we present our findings from training the LLaMA2-7B model on the OpenPlatypus and Alpaca datasets. The OpenPlatypus dataset \cite{lee2023platypus}, and the Alpaca dataset \cite{alpaca}, are both designed to enhance instruction-following capabilities in language models. To evaluate the performance of our method, we report the accuracy on the Massive Multitask Language Understanding (MMLU) benchmark \cite{hendrycksmeasuring}, a comprehensive suite designed to test models across a wide range of academic and professional subjects.

\paragraph{Results.}
Table \ref{tab:instruction-tune} summarizes our results. Our experiments reveal that RoSA does not consistently outperform LoRA on instruction-tuning, particularly when tuning on datasets such as OpenPlatypus and Alpaca, which contain data relatively similar to the pre-training data.

As discussed earlier in the paper (refer to Section \ref{sec:motiv}), the advantage of RoSA is more pronounced when the training data is rather complex, i.e. in settings where full fine-tuning significantly outperforms LoRA. This observation aligns with our current results, suggesting that for simpler instruction tuning tasks, LoRA performs adequately, matching or even outperforming FFT, and therefore RoSA is not necessary.

\paragraph{Analysis.}
The primary reason for RoSA's underperformance in these scenarios might be that, as mentioned earlier in the paper, when the tasks are not complex enough, RoSA's performance is on par with LoRA's. Another reason is based on the findings from \citet{he2022towards}, which indicate that added parameters are better utilized in the feed-forward network (FFN) layers rather than the attention layers. Since RoSA is more robust at capturing complex information, it is more effective when the added parameters are used in the MLP modules. Therefore, exploring different choices of target fine-tuning modules might be better to yield better performance; however, we leave this for further research.


\begin{table}[h]
\centering
\caption{\label{tab:instruction-tune} {MMLU 5-shot accuracy comparison on LLaMA2-7B. We observe that RoSA does not outperform LoRA on these simpler tasks.}}
\begin{tabular}{ccc}
\toprule
\textbf{Dataset}     &   \textbf{Model/Fine-tuning}    & \textbf{MMLU (5-shot)} \\ \hline
&\text{LLaMa2-7B}  & $45.75$ \\ \hline
\text{OpenPlatypus}&\texttt{LoRA} $r=16$ & $49.92$ \\
&\texttt{RoSA} $r=16,d=0.6\%$ & $46.54$ \\ \hline
\text{Alpaca}&\texttt{LoRA} $r=16$ & $45.8$ \\
&\texttt{RoSA} $r=16,d=0.6\%$ & $46.52$ \\ \toprule
\end{tabular}
\end{table}

\section{Qualitative Results}
\label{sec:qual}
Here we showcase some examples of the generated outputs on the GSM8k task, which includes mathematical reasoning. We consider the FFT model, as well as LoRA, SpA, and best RoSA models in the $40M$ budget (single epoch).

\begin{adjustwidth}{32pt}{32pt}
\textbf{Question:} \\
\texttt{John rents his car out 10 times a month for 3 hours each time.  He gets paid \$25 an hour. If his car payment is \$500, how much profit does he make on his car?}

\textbf{Answer (Ground Truth):} \\
\texttt{He rents his car 10*3=<<10*3=30>>30 hour a month \\
So he makes 25*30=\$<<25*30=750>>750 a month \\
That means he has a profit of 750-500=\$<<750-500=250>>250 a month \\
\#\#\#\# 250}

\textbf{Answer (FFT):} \\
\texttt{He rents his car out 10*3=<<10*3=30>>30 hours a month \\
So he makes 30*25=\$<<30*25=750>>750 a month \\
So his profit is 750-500=\$<<750-500=250>>250 \\
\#\#\#\# 250}

\textbf{Answer (LoRA):} \\
\texttt{He makes 3*10=\$<<3*10=30>>30 per month \\
So he makes 30*12=\$<<30*12=360>>360 per year \\
So he makes 360-500=\$<<360-500=140>>140 profit \\
\#\#\#\# 140}

\textbf{Answer (SpA):} \\
\texttt{He makes 3*10=\$<<3*10=30>>30 per month \\
So he makes 30*12=\$<<30*12=360>>360 per year \\
So he makes a profit of 360-500=\$<<360-500=140>>140 \\
\#\#\#\# 140}

\textbf{Answer (RoSA):} \\
\texttt{He makes 3*10=\$<<3*10=30>>30 per month \\
So he makes 30*25=\$<<30*25=750>>750 per month \\
So he makes 750-500=\$<<750-500=250>>250 per month \\
\#\#\#\# 250}
\end{adjustwidth}

In the above example, while both LoRA and SpA decide to calculate yearly profit, RoSA successfully solves the problem correctly.

\begin{adjustwidth}{32pt}{32pt}
\textbf{Question:} \\
\texttt{Mary is two years younger than Joan, who is five years older than Jessa. If Jessa is 20 years old, what is the sum of the ages of the three girls?}

\textbf{Answer (Ground Truth):} \\
\texttt{Joan is 20 + 5 = <<20+5=25>>25 years old.\\
Mary is 25 – 2 = <<25-2=23>>23 years old.\\
Therefore, the sum of their ages is 20 + 25 + 23 = <<20+25+23=68>>68.\\
\#\#\#\# 68}

\textbf{Answer (FFT):} \\
\texttt{Joan is 20 + 5 = <<20+5=25>>25 years old.\\
Mary is 25 - 2 = <<25-2=23>>23 years old.\\
The sum of the ages of the three girls is 23 + 20 + 25 = <<23+20+25=68>>68 years old.\\
\#\#\#\# 68}

\textbf{Answer (LoRA):} \\
\texttt{The sum of the ages of the three girls is 20 + 5 + 2 = <<20+5+2=27>>27 years.\\
\#\#\#\# 27}

\textbf{Answer (SpA):} \\
\texttt{Mary is 20 - 5 = <<20-5=15>>15 years old. \\
The sum of the ages of the three girls is 20 + 15 + 15 = <<20+15+15=50>>50 years. \\
\#\#\#\# 50}

\textbf{Answer (RoSA):} \\
\texttt{Joan is 20 + 5 = <<20+5=25>>25 years old.\\
Mary is 25 - 2 = <<25-2=23>>23 years old.\\
The sum of the ages of the three girls is 23 + 20 + 20 = <<23+20+20=63>>63 years.\\
\#\#\#\# 63}
\end{adjustwidth}

While all adaptation methods (including RoSA) fail to answer the question correctly, we see that LoRA and SpA completely fail to even process it. In contrast, RoSA calculates the ages correctly and only fails to sum them up at the end.

\newpage

\end{document}